\documentclass{article}

\usepackage{times}
\usepackage[pdftex]{graphicx}
\usepackage{amsmath}
\usepackage{amssymb}
\usepackage{authblk}
\usepackage{bm}
\usepackage{tabularx}
\usepackage{multirow}
\usepackage{algorithm}
\usepackage{algorithmic}

\usepackage{fmt}
\usepackage{mcr}
\usepackage{pkg}
\usepackage{thmE}
\usepackage{algE}
\usepackage{textcomp}

\usepackage[pagebackref=true,breaklinks=true,letterpaper=true,bookmarks=false]{hyperref}
\usepackage[margin=1in]{geometry}

\newcommand{\argmin}{\mathop{\rm argmin}\limits}

\begin{document}

\title{Case-based Similar Image Retrieval for Weakly Annotated Large Histopathological Images of Malignant Lymphoma Using Deep Metric Learning}

\author[1]{Noriaki Hashimoto}
\affil[1]{RIKEN Center for Advanced Intelligence Project}
\author[2]{Yusuke Takagi}
\affil[2]{Department of Computer Science, Nagoya Institute of Technology}
\author[2]{Hiroki Masuda}
\author[3]{Hiroaki Miyoshi}
\affil[3]{Department of Pathology, Kurume University School of Medicine}
\author[3]{Kei Kohno}
\author[3]{Miharu Nagaishi}
\author[3]{Kensaku Sato}
\author[3]{Mai Takeuchi}
\author[3]{Takuya Furuta}
\author[3]{Keisuke Kawamoto}
\author[3]{Kyohei Yamada}
\author[3]{Mayuko Moritsubo}
\author[3]{Kanako Inoue}
\author[3]{Yasumasa Shimasaki}
\author[3]{Yusuke Ogura}
\author[3]{Teppei Imamoto}
\author[3]{Tatsuzo Mishina}
\author[3]{Ken Tanaka}
\author[3]{Yoshino Kawaguchi}
\author[4]{Shigeo Nakagmura}
\affil[4]{Department of Pathology and Laboratory Medicine, Nagoya University Hospital}
\author[3]{Koichi Ohshima}
\author[2]{Hidekata Hontani}
\author[1,5,*]{Ichiro Takeuchi}
\affil[5]{Department of Mechanical Systems Engineering, Nagoya University\protect\\{\it  noriaki.hashimoto.jv@riken.jp, ichiro.takeuchi@mae.nagoya-u.ac.jp}}

\date{}
\maketitle

\begin{abstract}
  In the present study, we propose a novel case-based similar image retrieval (SIR) method for hematoxylin and eosin (H\&E)-stained histopathological images of malignant lymphoma.
  When a whole slide image (WSI) is used as an input query, it is desirable to be able to retrieve similar cases by focusing on image patches in pathologically important regions such as tumor cells.
  To address this problem, we employ attention-based multiple instance learning, which enables us to focus on tumor-specific regions when the similarity between cases is computed.
  Moreover, we employ contrastive distance metric learning to incorporate immunohistochemical (IHC) staining patterns as useful supervised information for defining appropriate similarity between heterogeneous malignant lymphoma cases.
  In the experiment with 249 malignant lymphoma patients, we confirmed that the proposed method exhibited higher evaluation measures than the baseline case-based SIR methods.
  Furthermore, the subjective evaluation by pathologists revealed that our similarity measure using IHC staining patterns is appropriate for representing the similarity of H\&E-stained tissue images for malignant lymphoma.
\end{abstract}

\section{Introduction}
%
%
Malignant lymphoma is a group of blood malignancies with more than 70 subtypes~\cite{BB24571652}.
Because each subtype has a different treatment strategy and prognosis, it is crucially important to identify the correct subtype through pathological diagnosis.
Digital pathology based on WSIs is increasingly becoming popular, where a WSI contains an extremely large (e.g., $100,000 \times 100,000$ pixels) digital image of an entire specimen invasively extracted from a patient~\cite{el2019automated,miyoshi2020deep}.
Given the WSI of a new malignant lymphoma patient, the goal of this SIR task is to retrieve ``similar'' WSIs from the database of past malignant lymphoma cases, where we need to find an appropriate ``similarity'' metric that is useful for pathological diagnosis of malignant lymphoma.

In practice, hematopathologists perform diagnosis of malignant lymphoma in the following two stages.
In the first stage, hematoxylin and eosin (H\&E) stained tissue slides are analyzed to narrow down the list of potential subtypes and determine a combination of immunohistochemical (IHC) stains.
In the second stage, the subtype is identified by analyzing the expression patterns of tissue slides stained with several IHC antibodies selected in the first stage.
Note that the IHC staining pattern is the list of IHC antibodies selected in the first stage and the IHC expression pattern is the observation result indicating whether target cells in each IHC stained tissue specimen positively react in the second stage.
For inexperienced pathologists, narrowing down the subtypes and determining IHC staining patterns (combinations of IHC stains) in the first stage is a challenging and time-consuming task.
Given an H\&E stained tissue specimen as an input query, the proposed case-based SIR method can retrieve similar cases from a database of past cases along with their IHC staining patterns and subtypes which were diagnosed by experienced hematopathologists.
Such retrieved similar cases and their IHC patterns and subtypes will help to support the decision-making of pathologists in the first stage.
%
%
%

The proposed case-based SIR method is constructed by learning an appropriate similarity or distance metric that has the following two properties:
The first property is that the similarity between a query case and a retrieved case is determined based on the tumor regions in the two WSIs.
Because the WSI of a malignant lymphoma specimen contains both normal and tumor cells, the similarity should be determined based only on information in the tumor region.
However, it is difficult to know the region of a WSI that contains tumor cells, and the proposed method has to be trained using WSIs that have no annotations for the tumor region\footnote{
It is an extremely time-consuming task for pathologists to manually annotate the tumor regions in a large WSI, and it is almost impossible to conduct such annotations for hundreds of WSIs.
}.
To overcome this difficulty, we effectively incorporate attention-based multiple-instance learning (MIL)~\cite{ilse2018attention,hashimoto2020multi} into case-based SIR tasks.
Because malignant lymphoma subtypes are characterized by information in the tumor region only, our basic idea is that the attention region extracted through attention-based MIL for subtype classification can be regarded as the tumor region.

The second property is that the similarity is defined in such a way that cases with similar IHC staining patterns
can be
retrieved.
Several conventional SIR methods for digital pathology~\cite{peng2019multi,hegde2019similar,shi2018pairwise,yang2020deep,erfankhah2019heterogeneity,
kalra2020yottixel,zheng2019encoding,zheng2020diagnostic}
employed distance metric learning (DML), which is based on the relevance of labels where feature learning is performed such that the distance between two feature vectors from the same-labeled images is shorter.
Most of the existing SIR methods employ the coincidence of subtype labels~\cite{peng2019multi,hegde2019similar,shi2018pairwise,erfankhah2019heterogeneity,
kalra2020yottixel,zheng2019encoding,zheng2020diagnostic} or histologic structures~\cite{hegde2019similar,yang2020deep} as the similarity between two images compared in distance metric learning.
As malignant lymphoma cases are highly heterogeneous, different combinations of IHC stains are used even among cases with the same subtype labels.
Owing to such heterogeneity of IHC staining patterns, the coincidence of subtype would not be a suitable distance metric.
In this study, we address this problem by using the similarity of IHC staining patterns as the similarity between cases.
We regard the similarity as the continuous relevance index between two cases, which enables us to learn a metric that properly incorporates the similarity of IHC staining patterns through contrastive DML~\cite{chopra2005learning}.

\begin{figure*}[!t]
\begin{center}
\includegraphics[width=0.8\linewidth]{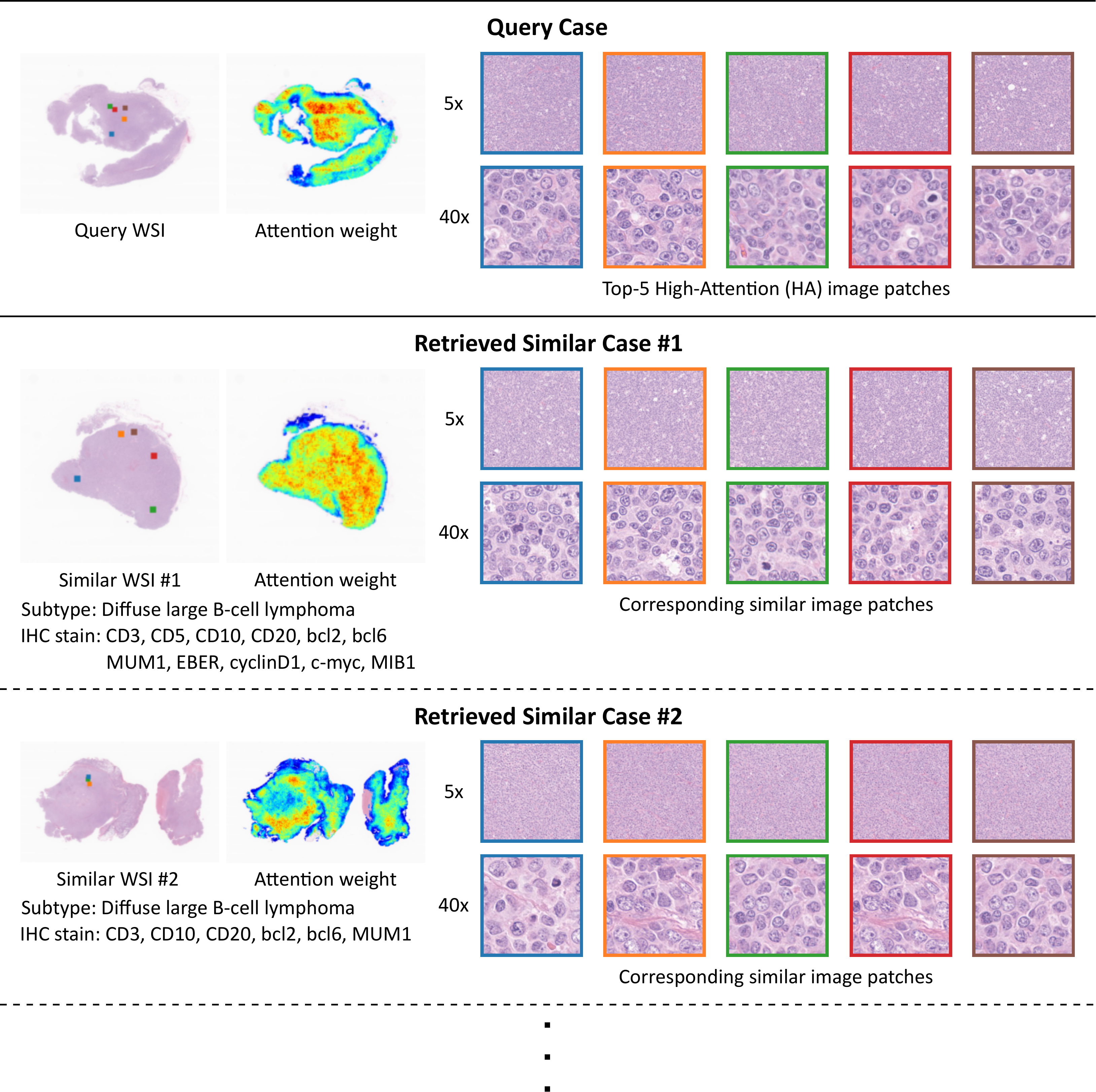}
\end{center}
\caption{
Example of the output of the proposed case-based SIR method.
When a user inputs a query case, the proposed method retrieves multiple similar past cases in the database.
The proposed method provides not only similar WSIs but also attention weight heatmaps and representative image patches that contribute to the determination of retrieved similar cases.
The five image patches for a query case are high-attention (HA) patches in which attention weights are the top-5 highest in the entire tissue, whereas the corresponding similar image patches for retrieved similar cases are the most similar image patches to each of the top-5 HA image patches.
The color frames in the image patches correspond to the small squares with same colors in the thumbnail of the WSI, indicating that each image patch was extracted from the corresponding region.
The main challenge addressed in this study is determining the attention weights and similarity between image patches for the pathological diagnosis of malignant lymphoma.
}
\label{fig:output}
\end{figure*}

Figure~\ref{fig:output} shows an example of the output of our case-based SIR method.
Because a WSI contains the entire specimen including normal regions, it is important to find cases in which there exist similar tumor regions, rather than finding cases in which the entire WSI is similar.
Therefore, as shown in Fig.~\ref{fig:output}, the proposed case-based SIR method not only provides similar past cases but also presents the tumor regions that are used to determine the similarity between a query and a retrieved similar case.
It helps pathologists to understand how the retrieved similar cases were selected, IHC staining patterns used by the experienced pathologists, and subtypes finally identified for the retrieved similar cases.
In this study, to verify the effectiveness of the proposed method, we applied it to 249 malignant lymphoma cases, each of which consisted of a WSI of a specimen, selected IHC staining patterns, and the final subtype diagnosed by experienced hematopathologists.
In addition to the quantitative evaluation based on the similarity of IHC staining patterns, subjective evaluations by 10 pathologists were conducted to confirm that the retrieved similar cases based on the obtained similarity metric are helpful for malignant lymphoma pathology.

\section{Preliminaries}
In this section, we first present the related works and our contributions.
Thereafter, we formulate the problem setup and evaluation measures of the case-based SIR method.
We also briefly describe the basic ideas of attention-based MIL and contrastive DML and explain the use of these techniques for developing the case-based SIR method.

\subsection{Related works and our contributions}
\paragraph{Related works}
In digital pathology, the availability of digital histopathological images has enabled numerous applications including
image classification~\cite{mousavi2015automated,hou2016patch,campanella2019clinical},
detection~\cite{cirecsan2013mitosis,cruz2014automatic,bejnordi2017diagnostic}, and
segmentation~\cite{xu2015deep,gao2016multi,tokunaga2019adaptive,tanizaki2020computing}.
Similar image retrieval is an important task in digital pathology~\cite{zheng2003design,caicedo2011content}, and recent advances in deep learning techniques have accelerated the development of SIR
methods~\cite{jimenez2017deep,komura2018luigi,schaer2019deep,peng2019multi,hegde2019similar,kalra2020yottixel}.

%
SIR methods in digital pathology are mainly categorized into two approaches: \emph{content-based SIR} and \emph{case-based SIR}.
In digital pathology, because a WSI is extremely large, analyses are usually performed on the basis of small image patches extracted from a WSI.
In content-based SIR, an image patch is given as a query, and similar image patches are retrieved as the results.
Examples of content-based SIR studies for digital pathology include \cite{zheng2019encoding} (lung cancer), \cite{zheng2018histopathological} (breast cancer), and \cite{peng2019multi} (colorectal cancer).
In contrast, case-based SIR methods must first select ``informative'' patches, and thereafter aggregate the similarities defined in multiple informative patches to form the overall WSI similarity.
In the context of digital pathology, relatively few studies on case-based SIRs have been conducted.
In \cite{jimenez2017deep}, an indicator known as the blue ratio, which is higher in regions with more cells, was used as the criterion for selecting informative patches.
In \cite{kalra2020yottixel}, clustering was first applied to patches, and a representative patch from each cluster was considered as an informative patch.
Although these approaches use simple methods for selecting informative patches, we employ attention-based MIL such that informative patches can be selected from the tumor region.
For pathological image classification based on WSIs that contain normal and tumor cells, MIL has been demonstrated to be effective~\cite{mercan2017multi,ilse2018attention,das2018multiple,couture2018multiple,wang2019rmdl,
campanella2019clinical,sudharshan2019multiple,hashimoto2020multi}.
Attention-based MIL~\cite{hashimoto2020multi} is particularly useful because it can quantify the relative importance of each image patch in the WSI as attention weight.
%
In the past works for digital pathology, it is known that high-attention regions in attention-based MIL correspond to subtype-specific regions and are often regarded as tumor regions because the information that determines subtype classification would exist in tumor cells~\cite{hashimoto2020multi,yao2020whole,lu2021data,hashimoto2022subtype}.
Actually, attention-based MIL was applied to the subtype classification of malignant lymphoma in which it was shown that computed attention regions corresponded to positive cells in the IHC stained tissue specimen~\cite{hashimoto2020multi}.
%
%
In this study, we incorporate attention-based MIL into case-based SIR tasks that enable the learning of distance measures based only on the information in the tumor region.

In SIR, DML has been effectively used to obtain an appropriate distance metric~\cite{wang2014learning,gordo2016deep,peng2019multi,hegde2019similar,shi2018pairwise,yang2020deep}.
DML methods can be roughly categorized into parametric distance measure- and feature learning-based approaches.
The former approach includes Mahalanobis DML~\cite{weinberger2009distance,shen2010scalable,chang2012boosting}
and multiple kernel learning~\cite{sonnenburg2006large,rakotomamonjy2007more,gonen2008localized,gonen2011multiple}.
When the SIR method is implemented with a deep neural network (DNN) model, a feature learning-based approach is often adopted.
For instance, given a class label for each image, feature learning is performed based on a loss function such that images with the same label are closer together, whereas images with different labels are farther apart~\cite{shi2018pairwise,hegde2019similar,peng2019multi}.
By selecting images based on the distance in the learned feature space, the distance metric in the SIR method can properly consider the class labels.
The proposed SIR method is designed to
retrieve similar cases with similar IHC staining patterns.
As mentioned, we regard the similarity of IHC staining patterns to the continuous similarity between the two cases.
Metric learning algorithms that use a continuous label or multi-label have been proposed~\cite{jin2009learning,gouk2016learning}.
We also define the continuous relevance index between two cases using IHC staining patterns and perform metric learning that utilizes pathological images.

\paragraph{Our contributions}
In this study, we propose a case-based SIR method that supports malignant lymphoma pathology diagnosis by introducing a DNN model that effectively incorporates MIL and DML.
To the best of our knowledge, there is no method that combines these techniques with the case-based SIR method.
Overall, the main advantages of the proposed method and our contributions in this study are summarized as follows:
\begin{itemize}
 \item
      By incorporating attention-based MIL into cased-based SIR tasks, the proposed method can retrieve similar cases based on a similarity measure that depends only on patches in the tumor region.

 \item
      By defining the similarity of two H\&E stained images using IHC staining patterns in contrastive DML, the proposed method can retrieve similar cases that would have similar IHC staining patterns.

 \item
      We applied the proposed method to 249 cases of malignant lymphoma and demonstrated its effectiveness through quantitative evaluation and subjective evaluation by 10 pathologists.
\end{itemize}

\subsection{Problem setup}
In this study, we denote the set of natural numbers up to $c$ as $[c] := \{1, \ldots, c\}$.
Let $N$ be the number of past malignant lymphoma cases (patients), $K$ be the number of subtypes, and $L$ be the number of the kinds of IHC stains.
The entire database of the past cases is represented as $\cT = \{(\mathbb{X}_n, \mathbb{Y}_n,\mathbb{S}_n)\}_{n \in [N]}$, where $\mathbb{X}_n$ is a WSI, $\mathbb{Y}_n$ is a $K$-dimensional one-hot vector for the subtype, and $\mathbb{S}_n$ is an $L$-dimensional binary vector for IHC staining patterns.
Here, the position of 1 in the one-hot vector $\mathbb{Y}_n$ indicates the subtype, whereas the values 1 and 0 in the binary vector $\mathbb{S}_n$ indicate whether or not the corresponding IHC stain was used for the pathological diagnosis.

\begin{figure*}[!t]
\begin{center}
\includegraphics[width=0.9\linewidth]{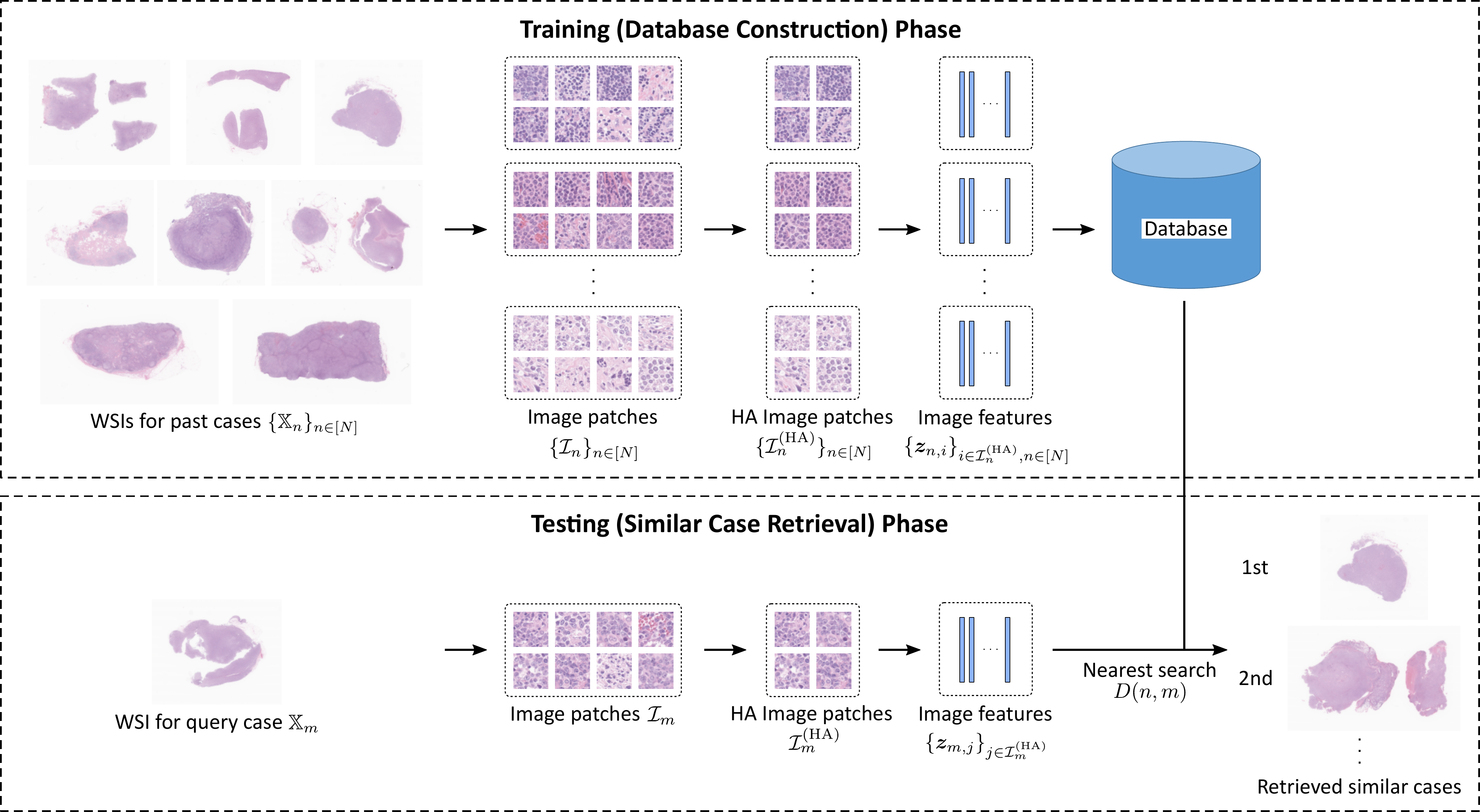}
\end{center}
\caption{
Overview of the training and test phases of the proposed case-based SIR model.
In the training phase, only HA image patches are sampled from the entire image patches of past cases, and the image features $\{\bm z_{n, i}\}_{i\in \cI_n^{\rm (HA)}}$ for HA patches $\{\cI_n^{\rm (HA)}\}_{n \in [N]}$ are saved in the search database.
In the test (similar case retrieval) phase, given a new query WSI $\mathbb X_m$, HA image patches $\cI_m^{\rm (HA)}$ are sampled similarly and their features $\{\bm z_{m, j}\}_{j\in \cI_m^{\rm (HA)}}$ are calculated.
By computing the case distance $D(n,m)$ between the query case $m$ and all the past cases $n \in [N]$, the case $n$ that has the minimum (resp., the 2nd, 3rd, $\ldots$ minimum) distance from a query case $m$ is retrieved as the most similar (resp., the 2nd, 3rd, $\ldots$ most similar) case.
}
\label{fig:SIR}
\end{figure*}

We develop a case-based SIR method using a database of past cases $\cT$, as shown in Fig.~\ref{fig:SIR}.
As discussed in Section 1, the proposed method is designed to
retrieve similar cases that would have similar IHC staining patterns using features in the tumor region only.

Consider a situation in which a WSI $\mathbb{X}_m$ of a new query case $m$ is input to the method, and we want to compute the distance between the query case $m$ and one of the past cases $n \in [N]$.
Let $\cI_n$ and $\cI_m$ be the sets of all image patches taken from ${\mathbb X}_n$ and ${\mathbb X}_m$, respectively, where $Q$ image patches are randomly sampled from the entire WSI\footnote{
We set $Q = 5000$ for the training phase and 1000 for the test phase in the demonstration study in Section 4.
}.
Furthermore, let $\cI_n^{\rm (HA)} \subset \cI_n$ and $\cI_m^{\rm (HA)} \subset \cI_m$ be the sets of image patches taken from the (estimated) tumor region in ${\mathbb X}_n$ and ${\mathbb X}_m$, respectively, where ``HA'' denotes ``High-Attention'' and we refer to those patches as \emph{HA patches}.
We denote the image patches from cases $n$ and $m$ as $\bm x_{n, i}, i \in \cI_n,$ and $\bm x_{m, j}, j \in \cI_m$, respectively.
Furthermore, let us denote the feature vectors (which will be learned through DNN representation learning whose details will be explained in Section 3) corresponding to image patches $\bm x_{n, i}$ and $\bm x_{m, j}$ as $\bm z_{n, i}$ and $\bm z_{m, j}$, respectively.
The desirable distance metric for the proposed case-based SIR method between query case $m$ and past case $n$ is defined as
\begin{align}
 \label{eq:prob}
 D(n, m)
 =
 \sum_{j \in \cI_m^{\rm (HA)}}
 \min_{i \in \cI_n^{\rm (HA)}}
 \|\bm z_{n, i} - \bm z_{m, j}\|_2.
\end{align}
Given a query WSI $\mathbb{X}_m$, the proposed case-based SIR method retrieves a (or a few) similar case $n$ such that the distance $D(n, m)$ is less than those of the remaining cases
\footnote{The pairs of five image patches in Fig.~\ref{fig:output} are the top-5 HA image patches of the query case and the corresponding similar image patches of the retrieved similar cases.
We display these pairs of image patches to explain the similarity between the query and retrieved similar cases.
}.

There are two challenges for learning the desirable distance metric Eq.~\eq{eq:prob}.
The first challenge is that the sets of patches in the tumor region $\cI_n^{\rm (HA)}$ and $\cI_m^{\rm (HA)}$ are unknown.
As mentioned in Section 1, to overcome this challenge, we employ attention-based MIL, the details of which will be described later.
The second challenge is learning the features $\bm z_{n, i}$ and $\bm z_{m, j}$ as a DNN representation such that the distances in Eq.~\eq{eq:prob} tend to be small when the IHC staining patterns ${\mathbb S}_n$ and ${\mathbb S}_m$ are similar.
As mentioned in Section 1, to overcome this challenge, we employ contrastive DML, the details of which will be described later.
In Section 3, we propose a DNN model and its learning algorithm by effectively combining attention-based MIL and contrastive DML for IHC staining patterns.

\subsection{Evaluation measure}

In the context of DML for classification problems, a common evaluation measure is simply the classification error (the subtype classification error in our problem setup).
However, the definitive diagnosis of lymphoma is determined by observing the IHC stained tissues and we consider that the subtype classification error is not sufficient as the evaluation measure for the similarity of H\&E stained tissue slides.
Because H\&E stained tissues of malignant lymphoma are highly heterogeneous, even cases with the same definitive subtype label have quite different IHC staining patterns.
Thus, the similarity of appearances of H\&E stained tissue specimens is reflected in the relevance of IHC staining patterns rather than the subtypes.
As a quantitative performance measure of the proposed case-based SIR method, we thus employ the Jaccard index for IHC staining patterns.
Given a query case $m$ and a retrieved case $n$, the Jaccard index of their IHC staining patterns ${\mathbb S}_m$ and ${\mathbb S}_n$ is defined as follows:
\begin{align}
 \label{eq:jaccard}
 r(n, m) := \frac{|{\mathbb S}_n \cap {\mathbb S}_m|}{|{\mathbb S}_n \cup {\mathbb S}_m|}.
\end{align}

For example, we consider the case wherein we use six types of IHC stains CD20, CD30, CD79a, bcl2, bcl6 and MUM1 (in this example, $L=6$) and the elements of the multi-label vector $\mathbb{S}_n$ correspond to [CD20, CD30, CD79a, bcl2, bcl6, MUM1].
Here, if the medical record for patient 1 indicates that CD20, bcl2, bcl6, and MUM1 were used in the diagnosis and the medical record for patient 2 indicates that CD20, CD79a, bcl2, bcl6, and MUM1 were used in the diagnosis, the multi-label vectors $\mathbb{S}_1$ and $\mathbb{S}_2$ are given as $[1,0,0,1,1,1]$ and $[1,0,1,1,1,1]$, respectively.
For these two cases, the similarity of IHC staining patterns is calculated as $r=0.8$.
In our experimental dataset, 26 types of IHC stains are considered as candidates $(L=26)$ in medical records, and 7.06 IHC stains are used for a case on average.
Examples of IHC staining patterns in practical cases are shown in Fig.~\ref{fig:output}.

We verified that the similarity of IHC staining patterns in the form of Eq.~\eq{eq:jaccard} is more meaningful measure than subtype classification error for case-based SIR in practical malignant lymphoma pathology by conducting subjective evaluation experiments by 10 pathologists.
In the subjective evaluation experiments, given a query case, we present a pair of retrieved similar cases, one of which is selected based on a distance measure trained to comply with the IHC staining patterns, whereas the other is selected using a distance measure trained to minimize subtype classification error, and pathologists answer which of the two cases are more similar to the query case.
The details of the subjective evaluations are presented in Section 4.

\subsection{Attention-based MIL}
Here, we describe the basic idea of attention-based MIL~\cite{ilse2018attention}, which is used as a component of the proposed DNN model in the next section, and its use in the proposed case-based SIR method.
In attention-based MIL, we define a \emph{bag} as a set of image patches randomly sampled from a WSI.
The basic idea of attention-based MIL is to assign \emph{an attention weight} to each image patch, which indicates the relative importance of each image patch within the bag.
Let $\cB_n$ be the set of bags in the case $n \in [N]$ and $\cJ_{n,b}$ be the set of image patches in the bag $b$ of case $n$.
Furthermore, we denote $a_{n,b,i}$ as the attention weight of the image patch $i \in \cJ_{n,b}$.
The attention weight $a_{n,b,i}$ takes a value in $[0, 1]$, and it is normalized such that the sum of the attention weights in each bag is one, that is, $\sum_{i \in \cJ_{n, b}} a_{n, b, i} = 1$.
In attention-based MIL, a classifier (for malignant lymphoma subtype classification) is trained with the importance weighting of each image patch by the attention weight, whereas the attention weights are also adaptively updated during the training process.
Because malignant lymphoma subtypes are characterized by information in the tumor region only, image patches that have high attention weights are considered to be taken from the tumor region.
In this study, we assume that at least $M$ image patches in each bag are sampled from the tumor region; thus, the collection of the top-$M$ image patches in all the bags $b \in \cB_n$ are considered as the set of high-attention image patches $\cI_n^{\rm (HA)}$\footnote{
We set $M = 10\%$ of all the image patches in a bag in the demonstration study in Section 4.
}.

\subsection{Contrastive DML}
Here, we describe the basic idea of contrastive DML~\cite{chopra2005learning}, which is used as a component of the proposed DNN model in the next section, and its use in the proposed case-based SIR method.
The goal of conventional feature learning-based DML is to learn a function that maps an image patch $\bm x_{n, i}$ to a feature vector $\bm z_{n, i}$ for $n \in [n], i \in \cI_n$ such that the Euclidean distance between the features $\bm z_{n, i}$ and $\bm z_{m, j}$ is small if the cases $n$ and $m$ belong to the same class, that is, ${\mathbb Y}_n = {\mathbb Y}_m$.
As discussed, because we intend to retrieve similar cases that would have similar IHC staining patterns rather than just belonging to the same subtype, we need to incorporate the similarity of IHC staining patterns into the distance metric.
Let $d(\bm x_{n, i}, \bm x_{m, j}) := \|\bm z_{n, i} - \bm z_{m, j}\|_2$, and consider the problem of learning the distance function $d(\cdot, \cdot)$ to minimize the following loss function:
\begin{align}
 \nonumber
 \sum_{(n, m) \in [N]^2 }
 &
 \sum_{(i, j) \in \cI_n^{\rm (HA)} \times \cI_m^{\rm (HA)}}
 \big\{
 r(n, m) d(\bm x_{n, i}, \bm x_{m, j})^2
 \\
 \label{eq:contrastive_loss}
 &
 + (1 - r(n, m)) {\rm max}\left(G - d(\bm x_{n, i}, \bm x_{m, j}), 0\right)^2
 \big\},
\end{align}
where $G$ is a hyperparameter that defines a margin between dissimilar image patches\footnote{
We set $G = 1.0$ in the demonstration study in Section 4.
}
(see Eq.~\eq{eq:concrete_distance_function} in Section 3.1 for the concrete formulation of the distance function $d(\cdot, \cdot)$).
In Eq.~\eq{eq:contrastive_loss}, $r(n, m)$ is known as the \emph{relevance index} in the context of contrastive DML, and we employ the Jaccard index in Eq.~\eq{eq:jaccard} as the relevance index for our task.
The first term in Eq.~\eq{eq:contrastive_loss} works such that image features of two inputs with similar labels are closer together, whereas the second term works such that image features of two inputs with different labels are farther apart based on the margin $G$.
By learning a feature extraction function that minimizes the loss in Eq.~\eq{eq:contrastive_loss}, we can obtain a distance function $d(\cdot, \cdot)$ that incorporates the similarity of the IHC staining patterns.

\section{Proposed case-based SIR method}
We propose a DNN model and its learning algorithm that provides the desirable distance metric in Eq.~\eq{eq:prob} for our case-based SIR task by effectively combining attention-based MIL and contrastive DML.
The problem of learning the desirable distance metric in Eq.~\eq{eq:prob} is decomposed into two sub-problems as follows:
The first sub-problem is to learn a function for extracting a set of image patches $\cI_n^{\rm (HA)}$ from the estimated tumor region.
The second sub-problem is to learn a function that maps an image patch $\bm x_{n, i}$ into a feature vector $\bm z_{n, i}$ for $n \in [N]$, $i \in \cI_n^{\rm (HA)}$, the latter of which is used to measure the distance in Eq.~\eq{eq:prob}.
Each of these two functions is obtained as a part of the entire DNN model.

\subsection{DNN model}
Figure~\ref{fig:model} illustrates the entire DNN model that consists of four components: $f_{\rm enc}$, $f_{\rm att}$, $f_{\rm clf}$, and $f_{\rm met}$, each of which is parameterized by a set of learnable parameters $\theta_{\rm enc}$, $\theta_{\rm att}$, $\theta_{\rm clf}$, and $\theta_{\rm met}$, respectively.
Each component is described as follows:

\begin{itemize}
 \item
      {\bf Feature extractor} $f_{\rm enc}$:
      The first component $f_{\rm enc}$ is known as the \emph{feature extractor}, which is introduced such that the two aforementioned sub-problems have common shared features.
      The feature extractor is a mapping as follows:
      \begin{align}
       f_{\rm enc}: \bm x_{n, i} \mapsto \bm h_{n, i}, i \in \cI_n, n \in [N],
      \end{align}
      where $\bm h_{n, i}$ denotes a feature vector of the image patch $\bm x_{n, i}$, which is implicitly defined by learning the representation in the DNN model.

 \item
      {\bf Attention network} $f_{\rm att}$:
      The second component $f_{\rm att}$ is used to compute the attention weights $a_{n, b, i}$, $n \in [N], b \in \cB_n, i \in \cJ_{n, b} \subset \cI_n$, and it is formally expressed as follows:
      \begin{align}
       f_{\rm att}: \{\bm h_{n, i}\}_{i \in \cJ_{n, b}} \mapsto \{a_{n, b, i}\}_{i \in \cJ_{n, b}}, n \in [N], b \in \cB_n.
      \end{align}
      Particularly, the attention weight is computed as follows:
      \begin{align}
       \label{eq:attention_weight}
       a_{n, b, i} = \frac{
       \exp
       \left(
       \bm{w}^\top {\rm tanh}(\bm{V} \bm{h}_{n, i})
       \right)
       }{
       \sum_{j \in \cJ_{n, b}}
       \exp
       \left(
       \bm{w}^\top {\rm tanh}(\bm{V} \bm{h}_{n, j})
       \right)
       },
       i \in \cJ_{n, b},
      \end{align}
      where $\bm V$ denotes a matrix of parameters, and $\bm w$ denotes a vector of parameters with appropriate dimensions, that is, $\theta_{\rm att} := (\bm{V}, \bm{w})$.

 \item
      {\bf Classifier network} $f_{\rm clf}$:
      The third component $f_{\rm clf}$ is used to classify the malignant lymphoma subtype based on the MIL framework (see Section 2.4).
      In MIL, a bag (a set of image patches randomly sampled from a WSI) $\cJ_{n, b}$, $b \in \cB_n, n \in [N]$, is classified into one of the $K$ subtypes.
      The input of $f_{\rm clf}$ is the weighted feature vector with attention weights as follows:
      \begin{align}
       \bm u_{n,b} := \sum_{i \in \cJ_{n, b}} a_{n,b,i} \bm h_{n,i}, n \in [N], b \in \cB_n.
      \end{align}
      Given an input $\bm u_{n,b}$, the subtype classifier outputs the $K$-dimensional class probability vector $P(\hat{\bm Y}_{n,b})$.
      Note that constructing a subtype classifier is not the main purpose of this study.
      By training the subtype classifier in the MIL framework, the attention network $f_{\rm att}$ is trained such that image patches taken from the tumor region have large attention weights.

 \item
      {\bf Metric network} $f_{\rm met}$:
      The fourth component $f_{\rm met}$ is used to transform the feature vector $\bm h_{n, i}$, $n \in [N]$, $i \in \cI_n^{\rm (HA)}$, obtained by the feature extractor $f_{\rm enc}$ into another feature vector $\bm z_{n, i}$, which is used for the desirable distance metric in Eq.~\eq{eq:prob} through contrastive DML (see Section 2.5).
      The metric network $f_{\rm met}$ is trained such that the contrastive loss in Eq.~\eq{eq:contrastive_loss} is minimized, where the distance function $d(\bm x_{n, i}, \bm x_{m, j})$ is implemented with $f_{\rm enc}$ and $f_{\rm met}$ as follows:
      \begin{align}
       \nonumber
       d(\bm x_{n, i}, \bm x_{m, j})
       &
       :=
       \|
       \bm z_{n, i}
       -
       \bm z_{m, j}
       \|_2
       \\
       \label{eq:concrete_distance_function}
       &
       =
       \|
       f_{\rm met}(f_{\rm enc}(\bm x_{n, i}))
       -
       f_{\rm met}(f_{\rm enc}(\bm x_{m, j}))
       \|_2.
      \end{align}
      Note that when $f_{\rm met}$ is trained, only parts of the image patches $\{\cI_n^{\rm (HA)}\}_{n \in [N]} \subset \{\cI_n\}_{n \in [N]}$ are used.
      As described in Section 2.4, $\cI_n^{\rm (HA)}$ is the set of image patches whose attention weights are within the top $M$ in each bag $b \in \cB_n$.
\end{itemize}

\subsection{Training DNN model}
\begin{figure*}[t]
  \begin{center}
      \includegraphics[width=0.8\textwidth]{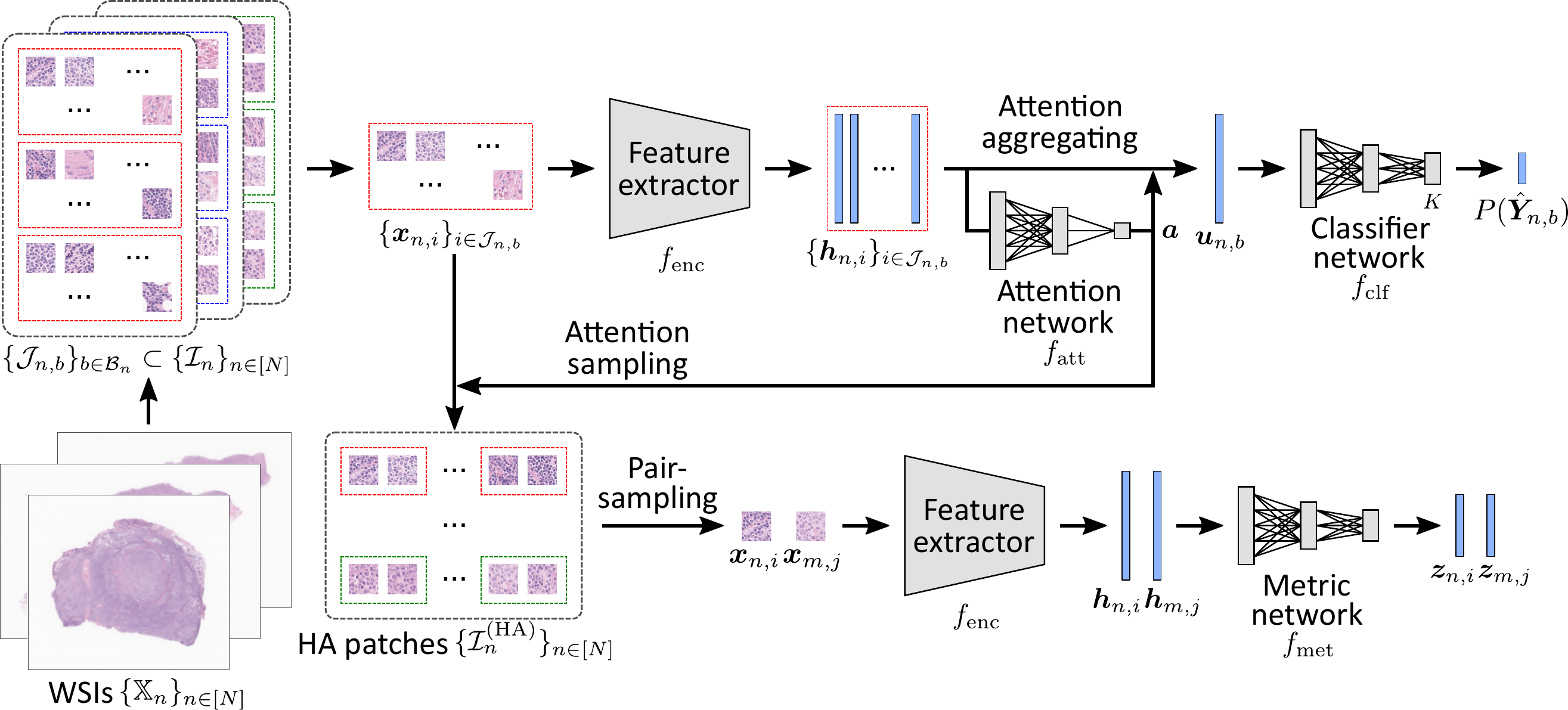}
\end{center}
  \caption{Schematic illustration of the DNN model for the proposed method.
The DNN model effectively combines attention-based MIL classification to identify tumor-specific image patches as HA patches and contrastive DML to learn the appropriate distance metric by incorporating the similarity of IHC staining patterns.
  The proposed DNN model consists of four components, $f_{\rm enc}$, $f_{\rm att}$, $f_{\rm clf}$, and $f_{\rm met}$, each of which is parameterized by a set of learnable parameters $\theta_{\rm enc}$, $\theta_{\rm att}$, $\theta_{\rm clf}$, and $\theta_{\rm met}$, respectively.
  In attention sampling, image patches, that had the top-$M$ highest attention weights in a bag, are extracted during the training of the MIL classification model.
Pair-sampling is a process that randomly selects a pair of two image patches from extracted HA image patches by attention sampling.
The contrastive DML is trained such that the loss function $\cL_d$ becomes minimum using the calculated distance between two HA image patches and the relevance of their IHC staining patterns.
  The parameters for attention-based MIL and contrastive DML are alternately updated.
  }
  \label{fig:model}
\end{figure*}

The parameters of the four components $\theta_{\rm enc}$, $\theta_{\rm att}$, $\theta_{\rm clf}$, and $\theta_{\rm met}$, respectively, for $f_{\rm enc}$, $f_{\rm att}$, $f_{\rm clf}$, and $f_{\rm met}$ are optimized by the alternate algorithm following two minimization problems:
\begin{align}
  \left(
  \hat{\theta}_{\rm enc},
  \hat{\theta}_{\rm att},
  \hat{\theta}_{\rm clf}
  \right)
   \leftarrow
  \argmin_{
  \theta_{\rm enc},
  \theta_{\rm att},
  \theta_{\rm clf}
  }
  \sum_{n \in [N]}
  \sum_{b \in \cB_n}
  \cL_{\rm c}(\mathbb{Y}_n, P(\hat{\bm Y}_{n, b})),
  \\
\left(
   \hat{\theta}_{\rm enc},
   \hat{\theta}_{\rm met}
   \right)
    \leftarrow
   \argmin_{
   \theta_{\rm enc},
   \theta_{\rm met}
   }
   \sum_{(n, m) \in [N]^2}
   \sum_{(i, j) \in \cI_n^{\rm (HA)} \times \cI_m^{\rm (HA)}}
   \mathcal{L}_{\rm d}(\bm{z}_{n, i},\bm{z}_{m, j}),
\end{align}
where
the loss function $\cL_{\rm c}$ is the standard cross-entropy loss defined as follows:
\begin{align}
 \cL_{\rm c}(\mathbb{Y}_n, P(\hat{\bm Y}_{n, b}))
 :=
 -
 \sum_{k \in [K]}
 \mathbb{Y}_{n, k}
 \log P(\hat{\bm Y}_{n, b})_k,
\end{align}
whereas the loss function $\cL_{\rm d}$ is the contrastive loss function (see Eq.~\eq{eq:contrastive_loss}), defined as follows:
\begin{align}
  \nonumber
 \cL_d(\bm z_{n, i}, \bm z_{n, j})
 &
 :=
 r(n, m)
 \|\bm z_{n, i} - \bm z_{m, j}\|_2^2
 \\
 &
 +
 (1 - r(n, m)) {\rm max}\left(G - \|\bm z_{n, i} - \bm z_{m, j}\|_2, 0\right)^2.
\end{align}
In Fig.~\ref{fig:model}, attention sampling extracts image patches, that had the top-$M$ highest attention weights in a bag, during the training of the MIL classification model.
Pair-sampling is a process in training the DML model, in which a pair of two image patches are randomly selected from HA image patches extracted by attention sampling.
The contrastive DML is trained such that the loss function $\cL_d$ becomes minimum using the calculated distance between two HA image patches sampled by pair-sampling and the relevance of their IHC staining patterns.
In our implementation, four components are implemented as follows:
We employ ResNet50~\cite{he2016deep} as the feature extractor $f_{\rm enc}$, and it is initialized with the extractor pre-trained with the ImageNet database~\cite{deng2009imagenet}.
The attention network $f_{\rm att}$ is implemented as a softmax operator in Eq.~\eq{eq:attention_weight}.
The classifier network $f_{\rm clf}$ is implemented using a simple neural network for $K$-class classification.
The metric network $f_{\rm met}$ is implemented using a simple neural network for feature transformation.
The implementation details are described in Section 4.1.

\subsection{Case-based SIR based on the trained DNN model}
For the construction of the search database in the case-based SIR task, $Q$ image patches are randomly sampled from each WSI in the training dataset, and the attention weights of these image patches are calculated by the trained feature extractor $f_{\rm enc}$ and the trained attention network $f_{\rm att}$.
Thereafter, from these image patches, $M$ image patches that have higher attention weights than other image patches are selected as the HA patches $\{\bm x_{n, i}\}_{i \in \cI_n^{\rm (HA)}}$, and their feature vectors $\{\bm z_{n, i}\}_{n \in [N], i \in \cI_n^{\rm (HA)}}$ are saved on the database as references.
Although the training phase of this model requires subtype labels and IHC staining patterns in addition to WSIs of patients to correctly train the attention network and the metric network, the testing (retrieval) phase requires only WSIs as an input to the system.
In the testing (retrieval) phase, when a WSI of a new query case $m$ is input to the SIR model, $M$ of HA image patches are sampled from $Q$ image patches $\{\bm x_{m, j}\}_{j \in \cI_m}$, and the feature vectors for HA image patches $\{\bm z_{m, j}\}_{j \in \cI_m^{\rm (HA)}}$ are computed using the aforementioned procedure.
%
%
For each $n \in [N]$, the distance $D(n, m)$ in Eq.~\eq{eq:prob} is calculated, and the most similar case (or multiple cases with the highest similarity) are retrieved.
Along with the selection of the most similar case(s), a set of similar image patch pairs is also provided as additional information (see Section 2.2 and Fig.~\ref{fig:output}).

\paragraph{Multi-scale input}

Because pathologists observe the H\&E stained tissue slides under a microscope at different magnifications, it is preferred that the retrieval results are also based on similarity using multi-scale information.
In the training phase using multi-scale inputs, different DNN models are independently trained with image patches of the corresponding magnifications.
In the testing (retrieval) phase using two magnifications (e.g., 40x and 5x), the distance between image patches of high and low magnifications is calculated similar to Eq.~\eq{eq:concrete_distance_function} as follows:
\begin{align}
 d^{({\rm H,L})}(\bm{x}^{({\rm H})}_{n,i},\bm{x}^{({\rm H})}_{m,j},
\bm{x}^{({\rm L})}_{n,i},\bm{x}^{({\rm L})}_{m,j}
 )
 :=
 ||\bm{z}^{({\rm H})}_{n,i}-\bm{z}^{({\rm H})}_{m,j}||_2
 +
 ||\bm{z}^{({\rm L})}_{n,i}-\bm{z}^{({\rm L})}_{m,j}||_2.
 \label{eq:multi_distance_function}
\end{align}
The embedded image features $\bm{z}_{n,i}^{({\rm H})}$ and $\bm{z}_{n,i}^{({\rm L})}$ are calculated from the image patches $\bm{x}_{n,i}^{({\rm H})}$ and $\bm{x}_{n,i}^{({\rm L})}$ for high and low magnifications, respectively.
Note that image patches $\bm{x}_{n,i}^{({\rm H})}$ and $\bm{x}_{n,i}^{({\rm L})}$ are extracted from the regions of the same central field of view in the WSI.
High-attention patches for multi-scale input are selected using the average attention weights for multiple magnifications $a^{({\rm H,L})}_{n,i}=(a^{({\rm H)}}_{n,i}+a^{({\rm L})}_{n,i})/2$.
Similar cases are obtained by comparing the case distances using multi-scale patch distances $d^{({\rm H,L})}(\bm{x}^{({\rm H})}_{n,i},\bm{x}^{({\rm H})}_{m,j},
\bm{x}^{({\rm L})}_{n,i},\bm{x}^{({\rm L})}_{m,j}
)$ between the HA image patches that are selected based on multi-scale attention weights $a^{({\rm H,L})}_{n,i}$.
If we employ three or more magnifications as the multi-scale input, the same number of DNN models are trained according to the increase in the number of input magnifications.

\section{Experiments}
To demonstrate the effectiveness of the proposed method, we applied it to 249 malignant lymphoma cases, each of which contains a WSI of a specimen, selected IHC stains, and the final subtype diagnosed by experienced hematopathologists.
In addition to the quantitative evaluation based on the similarity of IHC staining patterns, subjective evaluations by 10 pathologists were conducted to confirm that the retrieved similar cases based on the obtained similarity metric are useful in the pathological diagnosis of malignant lymphoma.

\subsection{Experimental setup}

\paragraph{Malignant lymphoma dataset}
The malignant lymphoma dataset contains $N = 249$ clinical cases with three subtypes: 76 diffuse large B-cell lymphoma (DLBCL), 90 follicular lymphoma (FL), and 83 reactive lymphoid hyperplasia (RL).
All cases were diagnosed at Kurume University in 2018, and the subtype labels for each case were identified by confirming the expression patterns of IHC stained tissue slides (called \emph{Kurume dataset} below).
For each case, we can refer to diagnostic information containing patient metadata, subtype, IHC staining pattern, and other findings.
In total, $L = 26$ types of IHC antibodies were used in this study.
All glass slides in the Kurume dataset were digitized using a WSI scanner Aperio GT 450 (Leica Biosystems, Germany) at 40x the original magnification (0.26 {\textmu}m/pixel).
In the experiment using the Kurume dataset, the 249 cases were split into five groups while maintaining the ratio of subtypes, and 5-fold cross-validation was performed.
In each fold of cross-validation, 25\% of the training cases were used as validation cases, which were used for selecting hyperparameters.

Furthermore, for the external validation, we used another lymphoma dataset whose slides were prepared at more than 80 institutions.
All cases were sent to and diagnosed at Nagoya University Hospital as the diagnostic consultation from 2003 to 2018 (called \emph{Nagoya dataset} below).
The Nagoya dataset contains 208 clinical cases with three subtypes same as the Kurume dataset: 98 DLBCL, 99 FL, and 11 RL.
Similar to the Kurume dataset, we can refer to diagnostic information of the Nagoya dataset containing patient metadata, subtype, IHC staining pattern, and other findings.
Although some diagnostic information of the Nagoya dataset shows the IHC antibodies that were not used in the Kurume dataset, only the same 26 types of IHC antibodies were used for the quantitative evaluation.
All glass slides in the Nagoya dataset were digitized using a WSI scanner Aperio ScanScopeXT (Leica Biosystems, Germany) at 20x the original magnification (0.50 {\textmu}m/pixel).

Using a WSI software OpenSlide~\cite{goode2013openslide}, the sets of image patches $\{\cI_n\}_{n \in [N]}$ were extracted from the tissue regions that were determined using Otsu method~\cite{otsu1979threshold} through saturation of HSV color space.
In the Kurume dataset, the average number of 40x image patches that were extracted from tissue regions of a single WSI was 16896, and it was equivalent to 24.7\% of entire WSIs.
In contrast, in the Nagoya dataset, the average number of 20x image patches that were extracted from tissue regions of a single WSI was 12238, and it was equivalent to 45.2\% of entire WSIs.
For digital pathology, it is extremely important to consider the varieties of the staining conditions and the difference in the WSI scanners.
In this study, we do not perform any pre-processing for them, because the tissue slides in the dataset of Kurume University were prepared by using the same scanner with the same condition in which there was less variety of the staining conditions.
However, for the practical application, those problems lead to loss of the performance and have to be solved.
As representative methods to solve such problems, a lot of works have been reported for stain normalization~\cite{zanjani2018stain,mahapatra2020structure} and domain adversarial learning~\cite{lafarge2017domain,hashimoto2020multi}.

All the samples used in this study were approved by the Ethics Review Committee of Kurume University, Nagoya University Hospital and RIKEN in accordance with the recommendations of the Declaration of Helsinki.

\paragraph{Implementation details}
The feature extractor $f_{\rm enc}$ was initialized by the ResNet50 pre-trained with the ImageNet database, and the dimension size of the output feature vector $\bm h_{n,i}, n \in [N], i \in \cI_n$ was set to 2048 after global average pooling layer.
The parameter $\theta_{\rm enc}$ for feature extractor $f_{\rm enc}$ was shared in the MIL classification model and the distance metric learning model.
Each of the three networks $f_{\rm att}$, $f_{\rm clf}$, and $f_{\rm met}$ was a three-layer neural network: the attention network $f_{\rm att}$ had 512 hidden units with hyperbolic tangent function as an activate function and an output unit, the classifier $f_{\rm clf}$ had 512 hidden units with ReLU function as an activate function and three output units, and the distance metric network $f_{\rm met}$ had 512 hidden units with ReLU function as an activate function and 64 output units.
For attention-based MIL, 5000 image patches with $224 \times 224$ pixels were extracted from each WSI and 50 bags, each of which contained 100 image patches, were used\footnote{If a WSI was excessively small to extract 5000 non-overlapping image patches, only a small number of available image patches were used.}.
We determined the maximum number of bags for each case so that entire slides can be covered to some extent because the malignant lymphoma in this study did not have very localized tumors.
After one-epoch training of attention-based MIL, that is, all bags were used for training once, the training of contrastive DML was conducted by using HA patches whose attention weights were ranked at the top $M = 10\%$ of all image patches in each bag, that is, $50 \times 100 \times 0.1 = 500$ HA patches.
In contrastive DML, 100 HA patches were randomly selected from the 500 HA patches for each case, and $N_{\rm tr} \times 100 / 2$ HA patch pairs were constructed such that each selected HA patch was used only for one of the pairs, where $N_{\rm tr}$ denotes the number of training cases.
The training of contrastive DML was conducted for 10 epochs after one-epoch training of the attention-based MIL.
This process was repeated 10 times in overall training, that is, attention-based MIL was trained for 10 epochs in total, whereas contrastive DML was trained for 100 epochs in total.
The parameters of the network were optimized using stochastic gradient descent (SGD) momentum~\cite{qian1999momentum}, where the learning rate, momentum, and weight decay were set to 1.25$\times 10^{-4}$, 0.9, and $10^{-4}$, respectively.

In the case-based SIR task after training, 1000 image patches were randomly extracted from each WSI, and 100 HA image patches with attention weights within the top 10\% in each case were used to compute the case distance $D(n,m)$.
Similar cases were retrieved based on the case distance, and the method provided retrieval results in descending (resp. ascending) order of similarity (resp. distance).
In our experiment, we considered high-magnification image input, low-magnification image input, and multi-scale input of high- and low-magnification images as the magnification of the input image patches (see Fig.~\ref{fig:output}).
We chose 40x/20x and 5x as high and low magnification because they are the magnitudes commonly used by pathologists; 40x/20x is used to examine the detailed shape of tumor cells and 5x is used to understand the overall histology of the specimen, respectively.
In a multi-scale setting, both high- and low-magnification image patches were extracted from the same regions, i.e., a center part of a low-magnification image patch had the same regions as the entire corresponding high-magnification image patch.
In our experimental setting, 200 image patches extracted from 100 positions were comprised in a multi-scale bag.

\paragraph{Baseline methods}
We compared the following five methods:
\begin{itemize}
 \item {\bf pre-trained ResNet50 + all patches},
 \item {\bf subtype-based metric + all patches},
 \item {\bf staining-based metric + all patches},
 \item {\bf subtype-based metric + HA patches}, and
 \item {\bf staining-based metric + HA patches (proposed method)}.
\end{itemize}
Here, ``all patches'' represents that attention-based MIL was not employed and image patches were randomly selected, whereas ``HA patches'' represents that attention-based MIL was used for selecting HA image patches.
In ``all patches'' setting, 1000 image patches were first randomly extracted from each WSI, and the training of contrastive DML was conducted for 100 epochs in which 100 image patches were randomly extracted from the 1000 image patches.
The first method ``pre-trained ResNet50 + all patches'' is a simple baseline in which neither attention-based MIL nor contrastive DML was used, and the distance between two cases was simply measured by the distances between two feature vectors obtained by a pre-trained ResNet50 with the ImageNet database without any fine-tuning.
Furthermore, ``subtype-based metric'' indicates that the relevance index $r(m, n)$ in Eq.~\eq{eq:contrastive_loss} for contrastive DML was defined as 1 if the subtypes of the two cases are the same and 0 otherwise, whereas ``staining-based metric'' indicates that the Jaccard index of IHC staining patterns in Eq.~\eq{eq:jaccard} was used as the relevance index.
By comparing the proposed method with the first four baseline methods, we demonstrate the effect of selecting HA image patches through attention-based MIL and the effect of considering the similarity of IHC staining patterns through contrastive DML.

\subsection{Results}
One of the main contributions of this study is the utilization of IHC staining patterns to provide a useful similarity measure for heterogeneous malignant lymphoma cases.
The performance of the proposed and baseline methods was evaluated not only through a quantitative evaluation but also through a subjective evaluation by 10 pathologists.
First, in the quantitative evaluation, the similarity of IHC staining patterns between a test query case and a retrieved similar case were compared among the methods.
In the subjective evaluation, we examined whether IHC staining similarity is a more appropriate measure than subtype similarity for the pathological diagnosis of malignant lymphoma.

\paragraph{Quantitative evaluation}
Using only the Kurume dataset, we evaluated the case-based SIR performance based on the similarity of the IHC staining patterns between an input query case and a retrieved similar case in the form of IHC staining accuracy defined by the Jaccard index in Eq.~\eq{eq:jaccard}.
Table~\ref{tab:accs} summarizes the results with three types of magnifications: 40x, 5x, and multi-scale of 40x \& 5x.
In the table, the average IHC staining accuracies of top-5 retrieved similar cases are listed.
The results demonstrate that the proposed method has the highest IHC staining accuracy among all the methods for all three types of magnifications.
The differences in the IHC staining accuracies between the proposed and four baseline methods are statistically significant at a significance level of 0.05, except for the 40x image input.
Although the accuracy looks low in the result, this is because of the problem setting of SIR.
Since the specific IHC staining pattern exists in the dataset, some query cases have no cases with the completely same IHC staining pattern in the search database.
In practice, when we compute the accuracy of IHC staining patterns between query cases and \emph{ideal} similar cases (whose IHC staining patterns are most similar to queries), the mean value of upper bound accuracy of IHC staining patterns was 0.948.

\begin{table*}[!t]
    \caption{Comparison of IHC staining accuracy between query cases and retrieved similar cases through 5-fold cross-validation with three types of magnifications.
    Each result shows the mean and standard error of the average accuracies of the top-5 retrieved similar cases for each query case.
    The proposed method achieved the best IHC staining accuracy in all types of magnifications.
    The differences between the proposed method and all the baseline methods are statistically significant at the 0.05 level in 5x and multi-scales of 40x \& 5x.
    Note that the upper bound accuracy of IHC staining patterns was 0.948 in this problem setting.
    }
  \begin{center}
    \begin{tabular}{lccc}
  \hline
  \multirow{2}{*}{Methods} & \multicolumn{3}{c}{Magnifications}\\
   & 40x & 5x & 40x \& 5x \\ \hline \hline
  \multicolumn{1}{l}{Pre-trained ResNet50 + all patches} & 0.602$\pm$0.010 & 0.601$\pm$0.010 & 0.609$\pm$0.010 \\
  \multicolumn{1}{l}{Subtype-based metric + all patches} & 0.618$\pm$0.010 & 0.641$\pm$0.011 & 0.645$\pm$0.011 \\
  \multicolumn{1}{l}{Staining-based metric + all patches} & 0.642$\pm$0.011 & 0.650$\pm$0.011 & 0.660$\pm$0.012 \\
  \multicolumn{1}{l}{Subtype-based metric + HA patches} & 0.633$\pm$0.010 & 0.643$\pm$0.011 & 0.653$\pm$0.010 \\
  \multicolumn{1}{l}{Staining-based metric + HA patches (proposed)} & \underline{\bf{0.647$\pm$0.011}} & \underline{\bf{0.669$\pm$0.011}} & \underline{\bf{0.676$\pm$0.011}} \\
  \hline
  \end{tabular}
\end{center}
\label{tab:accs}
\end{table*}

We also compared the similarity of malignant lymphoma subtypes between an input query case and a retrieved similar case in the form of subtype accuracy, which takes 1 if the two subtypes are the same and 0 otherwise.
Table~\ref{tab:accl} summarizes the subtype accuracy results in the same format as Table~\ref{tab:accs}.
Although the criterion employed in the proposed method is not directly related to the subtype accuracy measure, the proposed method achieved the best performance among the five methods in two of the three magnification settings.
In the magnification setting of 40x, ``subtype-based metric + HA patches'' achieved the best performance.
This is reasonable because the subtype-based metric is directly tailored to subtype accuracy.
In terms of the reason why the proposed method with ``staining-based metric'' was better or comparative to the method with ``subtype-based metric,'' we conjecture that a good representation for IHC staining patterns is also a good representation for the subtype because the difference in IHC staining patterns reflects the heterogeneity of malignant lymphoma subtypes.

\begin{table*}[!t]
    \caption{
    Comparison of subtype accuracy between query cases and retrieved similar cases through 5-fold cross-validation with three types of magnifications.
  Each result shows the mean and standard error of the average accuracies of the top-5 retrieved similar cases for each query case.
    The proposed method achieved the best subtype accuracy in two out of three magnifications and it was comparable in the remaining magnification.
    Note that the subtype accuracy is directly tailored to subtype-based metric.
    We conjecture that the proposed method with a staining-based metric had good performance in terms of not only IHC staining accuracy but also subtype accuracy because a good representation of IHC staining patterns is also a good representation of subtypes.
    }
  \begin{center}
    \begin{tabular}{lccc}
  \hline
  \multirow{2}{*}{Methods} & \multicolumn{3}{c}{Magnifications}\\
  & 40x & 5x & 40x \& 5x \\ \hline \hline
  \multicolumn{1}{l}{Pre-trained ResNet50 + all patches} & 0.633$\pm$0.018 & 0.625$\pm$0.017 & 0.654$\pm$0.017 \\
  \multicolumn{1}{l}{Subtype-based metric + all patches} & 0.673$\pm$0.019 & 0.714$\pm$0.020 & 0.736$\pm$0.020 \\
  \multicolumn{1}{l}{Staining-based metric + all patches} & 0.578$\pm$0.021 & 0.600$\pm$0.020 & 0.565$\pm$0.022 \\
  \multicolumn{1}{l}{Subtype-based metric + HA patches} & \underline{\bf{0.720$\pm$0.020}} & 0.737$\pm$0.021 & 0.774$\pm$0.020 \\
  \multicolumn{1}{l}{Staining-based metric + HA patches (proposed)} & 0.712$\pm$0.021 & \underline{\bf{0.770$\pm$0.020}} & \underline{\bf{0.783$\pm$0.019}} \\
  \hline
  \end{tabular}
\end{center}
\label{tab:accl}
\end{table*}

To confirm the validity of attention weights in the attention-based MIL, an expert hematopathologist performed a subjective evaluation to confirm that HA image patches correspond to subtype-specific regions, i.e. tumor-specific regions in DLBCL and FL cases.
In this experiment, only DLBCL and FL cases of the Kurume dataset were used because RL cases have no tumor cells in tissue specimens and pathologists cannot discriminate if a provided image patch is tumor-region or not.
First, we randomly selected 100 cases from all 166 DLBCL and FL cases.
For each of the selected 100 cases, multi-scale attention weights $a^{({\rm H,L})}_{n,i}$ in Section 3.3 were computed in the multi-scale model using 40x and 5x images, and then image patches that had the highest and the lowest attention weights in the case were obtained for the evaluation.
The shuffled 200 pairs of 40x and 5x image patches were shown to a pathologist and he evaluated if each pair of image patches included tumor cells of DLBCL or FL in them.
In the evaluation, only a case that contained tumor cells both in 40x and 5x image patches was regarded as a tumor region because the multi-scale attention weight was calculated by using integrated multi-scale information.
As the result, 63 image patches were evaluated as tumor regions of DLBCL or FL in all 100 high-attention image patches,
while only 24 image patches were evaluated as tumor regions in all 100 low-attention image patches.
We confirmed the significant difference between high- and low-attention image patches for the presence of tumor cells,
which can show the validity of attention weights in the attention-based MIL.

\paragraph{External validation}

As an external validation, we had a retrieval experiment in which the 208 lymphoma cases in Nagoya dataset were used as query cases.
In this experiment, all 249 cases in the Kurume dataset are used as the training data and the search database.
Because the original WSIs of the Nagoya dataset were scanned at 20x magnification,
20x image patches were extracted from WSIs of the Kurume dataset similarly.
In the experiment, accuracies of IHC staining patterns by pre-trained ResNet50, subtype-based metric, and staining-based metric (proposed) were compared and the results are summarized as shown in Table~\ref{tab:accn}.
In the table, the average IHC staining accuracies of top-5 retrieved similar cases are listed similarly to the results in Table~\ref{tab:accs}.
The results demonstrate that the proposed method has higher IHC staining accuracy even when the institutions, where tissue slides were prepared, were different between query cases and the search database.
Although the accuracy degraded compared to the previous experiment, it is caused by the difference in the characteristics of the dataset.
The IHC staining patterns in the Nagoya dataset include IHC stains that were selected by general pathologists,
whereas the IHC staining patterns in the Kurume dataset were selected only by expert hematopathologists.
The Nagoya dataset can have redundant IHC stains in medical records unlike the Kurume dataset, and the upper bound accuracy of IHC staining patterns of the Nagoya dataset was computed as 0.831 whereas that of the Kurume dataset was 0.948.
Other causes of reduced accuracy are the difference in the staining condition of H\&E stained tissues and the scanning hardware as discussed in a lot of works in digital pathology.
We did not consider them owing to the uniformity of the Kurume dataset, but currently, many methods to solve them are proposed such as the stain normalization~\cite{zanjani2018stain,mahapatra2020structure} and domain adversarial learning~\cite{lafarge2017domain,hashimoto2020multi}.
Multi-label classification is originally a difficult problem setting to achieve high accuracy.
However, the accuracies of the SIR methods in the external validation are still low and an increment of the accuracy is required for the clinical application as future work.

The above experiments could show the improvement of the accuracies of the IHC staining pattern.
The following subjective experiment verifies that the similarity of IHC staining patterns is appropriate for the similarity measure of the appearance of H\&E stained tissue images.

\begin{table*}[!t]
    \caption{Comparison of IHC staining accuracy between query cases and retrieved similar cases with three types of magnifications.
    Each result shows the mean and standard error of the average accuracies of the top-5 retrieved similar cases for each query case.
    The proposed method achieved the best IHC staining accuracy in all types of magnifications.
    The differences between the proposed method and all the baseline methods are statistically significant at the 0.05 level in all magnification settings.
    Note that the upper bound accuracy of IHC staining patterns was 0.831 in this problem setting.
    }
  \begin{center}
    \begin{tabular}{lccc}
  \hline
  \multirow{2}{*}{Methods} & \multicolumn{3}{c}{Magnifications}\\
   & 20x & 5x & 20x \& 5x \\ \hline \hline
  \multicolumn{1}{l}{Pre-trained ResNet50 + all patches} & 0.540$\pm$0.008 & 0.529$\pm$0.008 & 0.546$\pm$0.009 \\
  \multicolumn{1}{l}{Subtype-based metric + HA patches} & 0.519$\pm$0.008 & 0.530$\pm$0.008 & 0.531$\pm$0.009 \\
  \multicolumn{1}{l}{Staining-based metric + HA patches (proposed)} & \underline{\bf{0.561$\pm$0.009}} & \underline{\bf{0.543$\pm$0.009}} & \underline{\bf{0.565$\pm$0.009}} \\
  \hline
  \end{tabular}
\end{center}
\label{tab:accn}
\end{table*}

\begin{figure*}[!t]
\begin{center}
  \fbox{
   \includegraphics[width=0.8\linewidth]{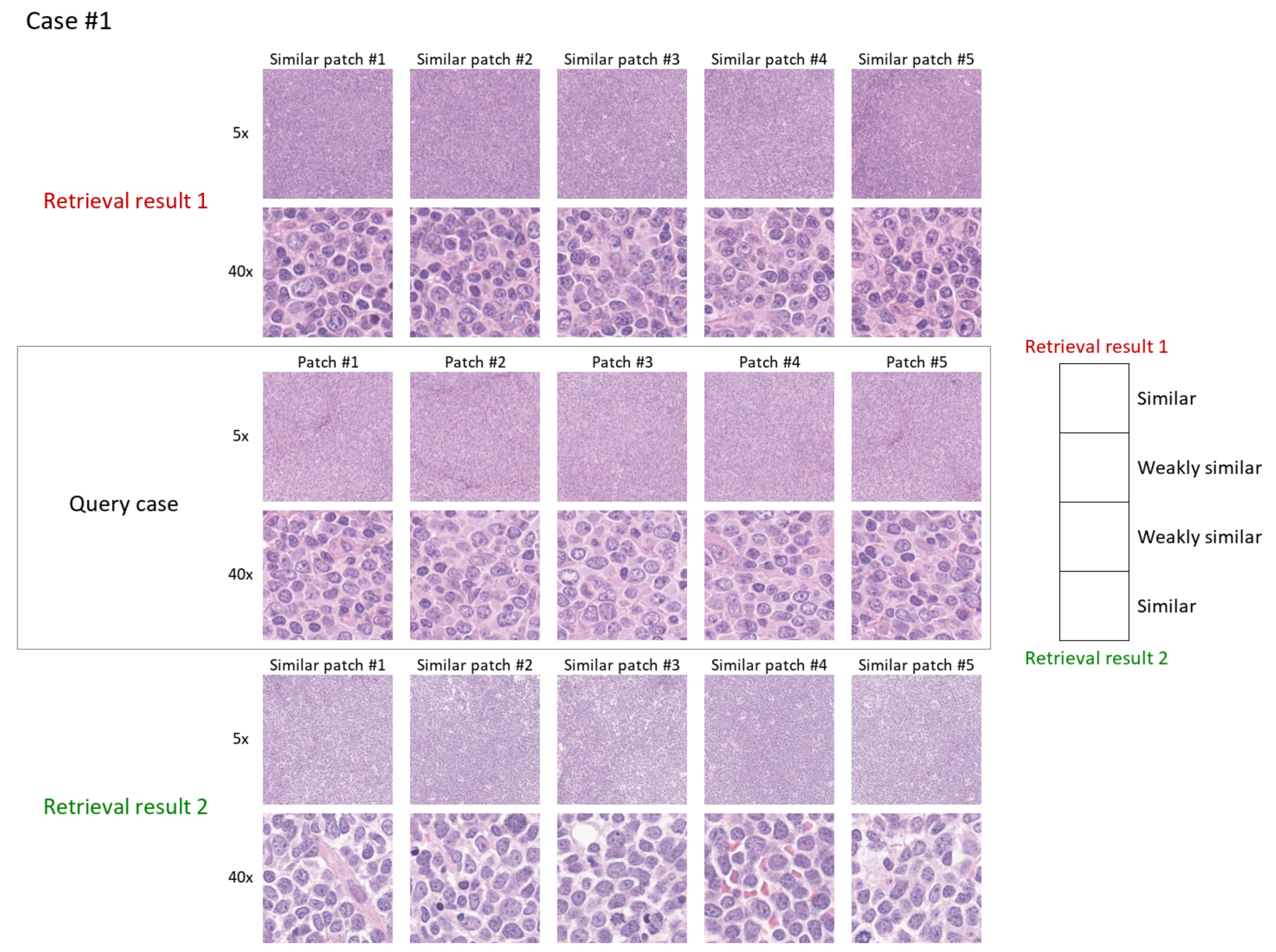}
   }
\end{center}
 \caption{
Example of the subjective evaluation tasks.
The participants were asked to evaluate the result that was more similar to a query case by a 4-grade score.
Five image patches of 40x and 5x magnifications that had top-5 attention weights were shown for an input query case, whereas the image patches that had the minimum distance from each query image patch obtained using the two methods are shown.
Either ``Retrieval result 1'' or ``Retrieval result 2'' corresponds to either the proposed method or the baseline method, which is determined at random.
}
\label{fig:sbj}
\end{figure*}

\paragraph{Subjective evaluation}
The goal of the subjective evaluation is to confirm whether IHC staining similarity is a more appropriate measure than subtype similarity for the pathological diagnosis of malignant lymphoma.
To this end, we only compared the proposed method ``{\bf staining-based metric + HA patches}'' with one of the baseline methods ``{\bf subtype-based metric + HA patches}'' in the subjective evaluations.
The task of each participant (pathologist) was to evaluate which of the two retrieval results (obtained using the proposed and baseline methods, respectively) was more similar to an input query.
An example of the subjective evaluation task is shown in Fig.~\ref{fig:sbj}.
For an input query case, the image patches of 40x and 5x magnifications with the top-5 multi-scale attention weights were shown, where attention weights were calculated in the three-class MIL classification model that was trained independently of the proposed and the compared SIR models.
The above multi-scale attentions $a_{n,i}^{\rm {(H,L)}}$ were calculated as written in Section 3.3 and a pair of multi-scale image patches extracted from the same region were shown.
For a retrieved similar case, the image patches that had the minimum distance from each query image patch obtained using each method were shown.
For instance, ``Similar patch \#1'' shows the most similar image patches corresponding to ``Patch \#1'' in the query case, where the pairs of image patches had the minimum distance $d^{({\rm H,L})}$ in Eq.~\eq{eq:multi_distance_function}.
All 249 cases were used as an input query once, that is, for each input query, the task was to find similar cases from the training (+validation) set for the cross-validation round when the input query was in the test set.
The result for each query was evaluated by a 4-grade score; a participant is asked to select one option among the following options: ``the result 1 is similar to a query,'' ``the result 1 is weakly similar to a query,'' ``the result 2 is weakly similar to a query'' or ``the result 2 is similar to a query,'' where either result 1 or result 2 corresponds to either the proposed method or the baseline method, which is determined at random and shown in blind.
In the experiment, all participants were asked to evaluate each case considering the similarity of both 40x and 5x image patches.
Note that this is the relative evaluation where the similarity of the result 1 indicates the dissimilarity of the result 2.
The order of query cases was also shuffled randomly for each participant.
In total, 10 pathologists composed of three experienced hematopathologists, three standard pathologists, and four pathological trainees participated in the subjective evaluation.

\begin{figure*}[t]
\begin{center}
   \includegraphics[width=0.85\linewidth]{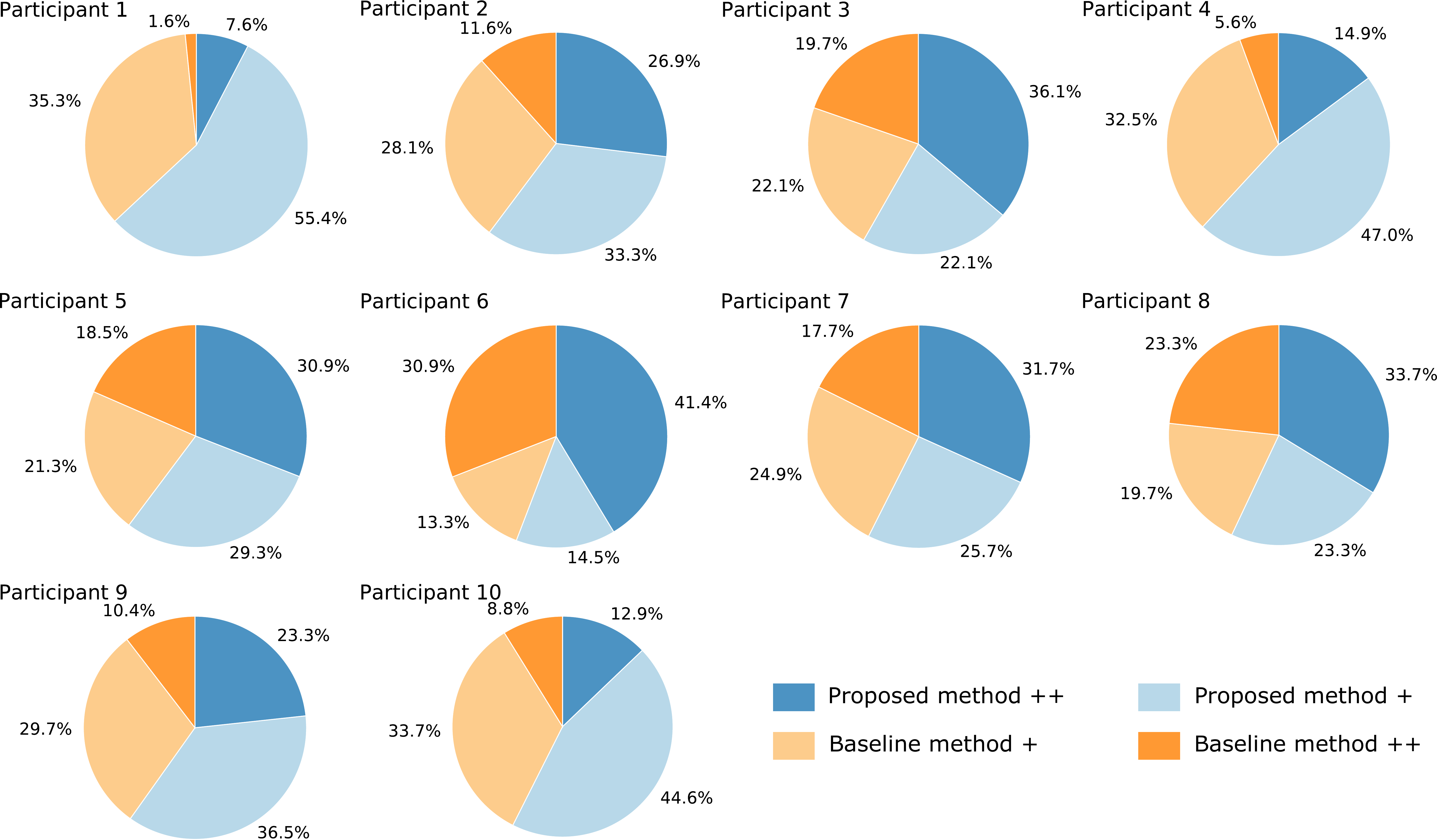}
\end{center}
 \caption{
Pie charts for the proportions of the four answers by 10 participants.
Each chart corresponds to a different participant, where ``++'' and ``+'' in legends mean ``similar to a query'' and ``weakly similar to a query,'' respectively.
It can be confirmed that thick blue and thin blue area are clearly more than a half in all participants, which indicates that the results retrieved using the proposed method were more likely to be evaluated as ``more similar.''
}
\label{fig:pie}
\end{figure*}

Figure~\ref{fig:pie} shows the results of each of the 10 participants in pie charts.
For all 10 participants, the proportion of responses in which the proposed method was more similar (thick blue) or weakly similar (thin blue) to the query case than the baseline method was significantly higher than the opposite responses (thick and thin orange colors).
This result indicates that all 10 pathologists determined that IHC staining similarity was more appropriate than subtype similarity as a similarity measure for the pathological diagnosis of malignant lymphoma.

To aggregate the evaluation results, evaluation score was counted as 1 if the result of the proposed method was evaluated as ``similar'' or ``weakly similar,'' and 0 otherwise.
We further compared ``confident responses'' by removing ``weakly similar'' responses.
Table~\ref{tab:sbj} lists the average evaluation scores of each of the 10 participants.
The results demonstrate that the proposed method could retrieve more similar cases in which pathologists felt they were more similar to query cases.
The superiority of the proposed method is more evident when we consider only the confident responses.
%
%
%
%
In all the presented results, the difference between the proposed and baseline methods is statistically significant with $p < 0.05$ based on a randomized test\footnote{
To quantify the statistical significance of the results in the subjective evaluation, we performed a Monte Carlo statistical test with the null hypothesis that the proposed and baseline methods are same.
Particularly, we generated 1,000,000 randomized results based on the null hypothesis.
A $p$-value for each participant is listed in Table~\ref{tab:sbj}.
Consequently, most of the 1,000,000 results are less than the actual scores listed in Table~\ref{tab:sbj}, and all the scores listed in Table~\ref{tab:sbj} are statistically significantly larger than 0.5, with $p < 0.05$.
}.
These results on subjective evaluation demonstrate that IHC staining similarity is more appropriate than subtype similarity for the pathological diagnosis of malignant lymphoma.
%

\begin{table}[!t]
    \caption{
    Mean binary scores of all participants through subjective evaluation experiment.
The ``Score'' indicates the results for all cases, whereas the ``Confident Score'' indicates the results answered with confidence, i.e., by excluding ``weakly similar'' responses.
    Each evaluation score was counted as 1 if the result of the proposed method was evaluated as ``similar'' or ``weakly similar,'' and 0 otherwise.
    The bracketed numbers indicate the number of ``confident responses'' by removing ``weakly similar'' responses.
The $p$-value for each result was computed by the Monte Carlo statistical test with the null hypothesis that the proposed and baseline methods are same, indicating that all the scores are highly statistically significant.
}
\begin{center}
  \begin{tabular}{ccccc}
\hline
Participants & Score & \multicolumn{2}{c}{Confident Score} & $p$-value \\ \hline \hline
\multicolumn{1}{c}{1} & 0.631 & ~~~0.826 & (23)~~~ & $2.8\times 10^{-5}$\\
\multicolumn{1}{c}{2} & 0.602 & ~~~0.698 & (96)~~~ & $7.6\times 10^{-4}$\\
\multicolumn{1}{c}{3} & 0.582 & ~~~0.647 & (139)~~~ & $5.6\times 10^{-3}$\\
\multicolumn{1}{c}{4} & 0.618 & ~~~0.725 & (51)~~~ & $1.2\times 10^{-4}$\\
\multicolumn{1}{c}{5} & 0.602 & ~~~0.626 & (123)~~~ & $7.6\times 10^{-4}$\\
\multicolumn{1}{c}{6} & 0.558 & ~~~0.572 & (180)~~~ & $3.8\times 10^{-2}$\\
\multicolumn{1}{c}{7} & 0.574 & ~~~0.642 & (123)~~~ & $1.1\times 10^{-2} $\\
\multicolumn{1}{c}{8} & 0.570 & ~~~0.592 & (142)~~~ & $1.5\times 10^{-2} $\\
\multicolumn{1}{c}{9} & 0.598 & ~~~0.690 & (84)~~~ & $1.2\times 10^{-3}$\\
\multicolumn{1}{c}{10} & 0.574 & ~~~0.593 & (54)~~~ & $1.1\times 10^{-2}$\\
\hline
\multicolumn{1}{c}{Mean$\pm$S.E.} & 0.591$\pm$0.0069 & \multicolumn{2}{c}{0.661$\pm$0.023} & \\
\hline
\end{tabular}
\end{center}
\label{tab:sbj}
\end{table}

\paragraph{Visualization of attention regions}
In case-based SIR, it is desirable to be able to explain the selection of retrieval results as similar cases.
To realize such explainable retrieval results, our proposed method provides the attention weights that indicate the regions that were focused as HA image patches in computing case distance $D(n,m)$.
The color plots of all WSIs in Fig.~\ref{fig:output} show the attention weights.
When we compute the attention weights for visualization purposes, attention weights $a_{n,i}$ or $a_{n,i}^{\rm (H,L)}$ for all image patches $i \in \cI_n$ were computed and normalized to the range $[0, 1]$ for each case $n \in [N]$.
The red region in the heat map indicates HA image patches that were used to calculate the similarity between two cases, whereas the blue region in the heat map indicates image patches whose attention weights are low.
In Fig~\ref{fig:output}, similar cases were retrieved with a multi-scale input of 40x \& 5x, and multi-scale attention weights $a_{n,i}^{\rm (H,L)}$ are visualized as a heat map.
We observe that the selected patches are visually similar; in particular, they are quite similar in both low and high magnifications by considering the multi-scale input.

\subsection{Examples}

\begin{figure}[t]
\begin{center}
   \includegraphics[width=0.5\linewidth]{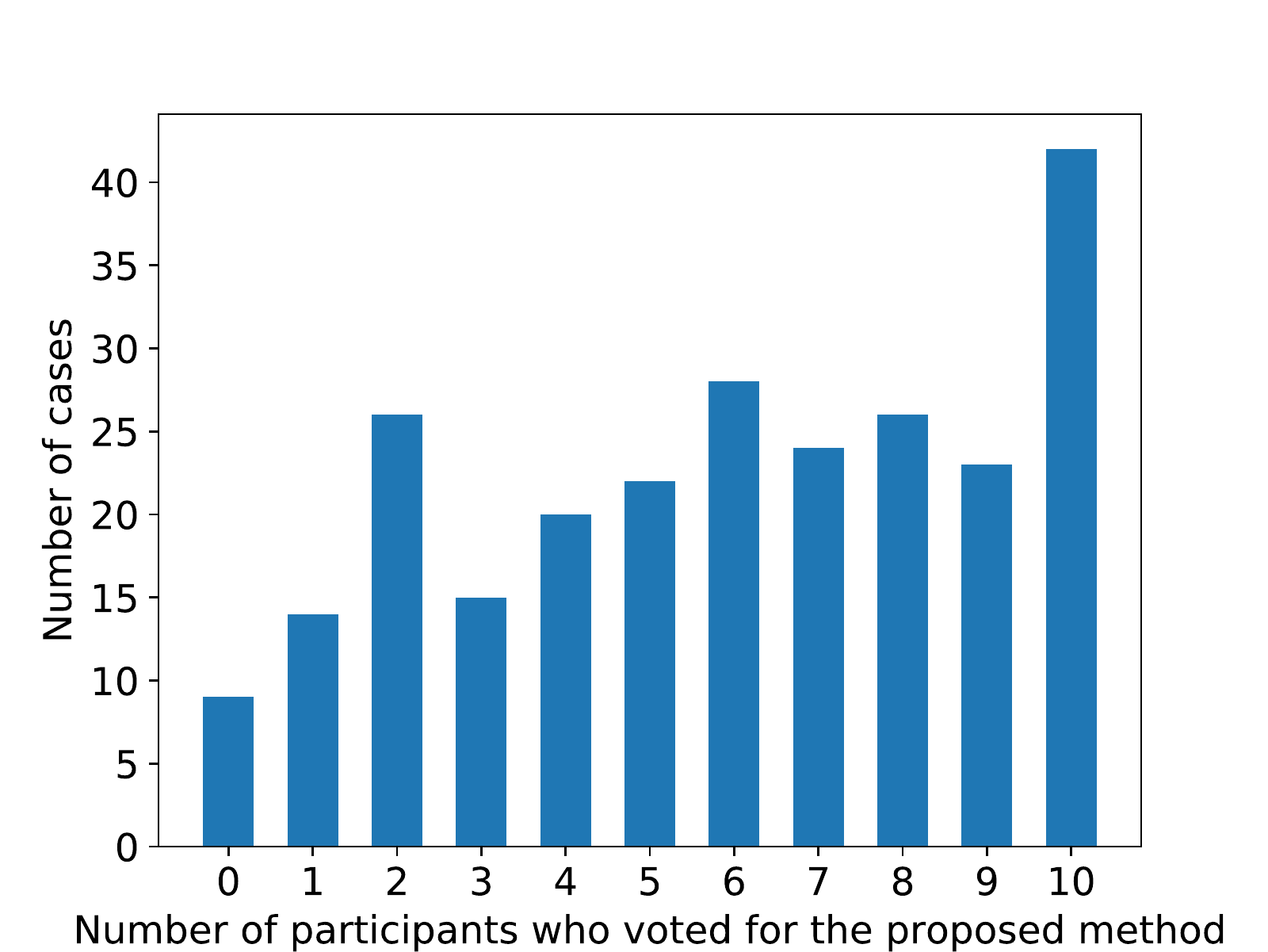}
\end{center}
 \caption{
Histogram of the number of cases for which the number of the 10 participants responded that the proposed method is better than the baseline method.
In total, in 42 cases, the proposed method was evaluated as more similar by all 10 participants, whereas there were only nine cases in which the baseline method was evaluated as more similar by all 10 participants.
}
\label{fig:bar}
\end{figure}

In the previously described experiments, we confirmed that the proposed method performed better than the baseline methods.
We investigate the results of the proposed method that were evaluated as more similar, and the difference of the results of the proposed method and those of the baseline method.
Figure~\ref{fig:bar} shows a histogram of the number of cases for which how many of the 10 participants responded that the proposed method is better than the baseline method in the subjective evaluation in Section 4.2.
The horizontal axis represents the number of participants who voted for the proposed method as the more similar result, e.g., ``10'' shows the number of cases in which all participants voted for the proposed method as ``similar'' or ``weakly similar.''
In these aggregated results, 143-case results of the proposed method were evaluated as more suitable than the baseline method by the majority of the participants.
In total, in 42 cases, the proposed method was evaluated as more similar by all 10 participants, whereas there were only nine cases in which the baseline method was evaluated as more similar by all 10 participants.

\begin{figure*}[t]
  \begin{center}
  \begin{tabular}{cc}
    \begin{minipage}[b]{0.40\hsize}
      \centering
        \fbox{
      \includegraphics[width=\hsize]{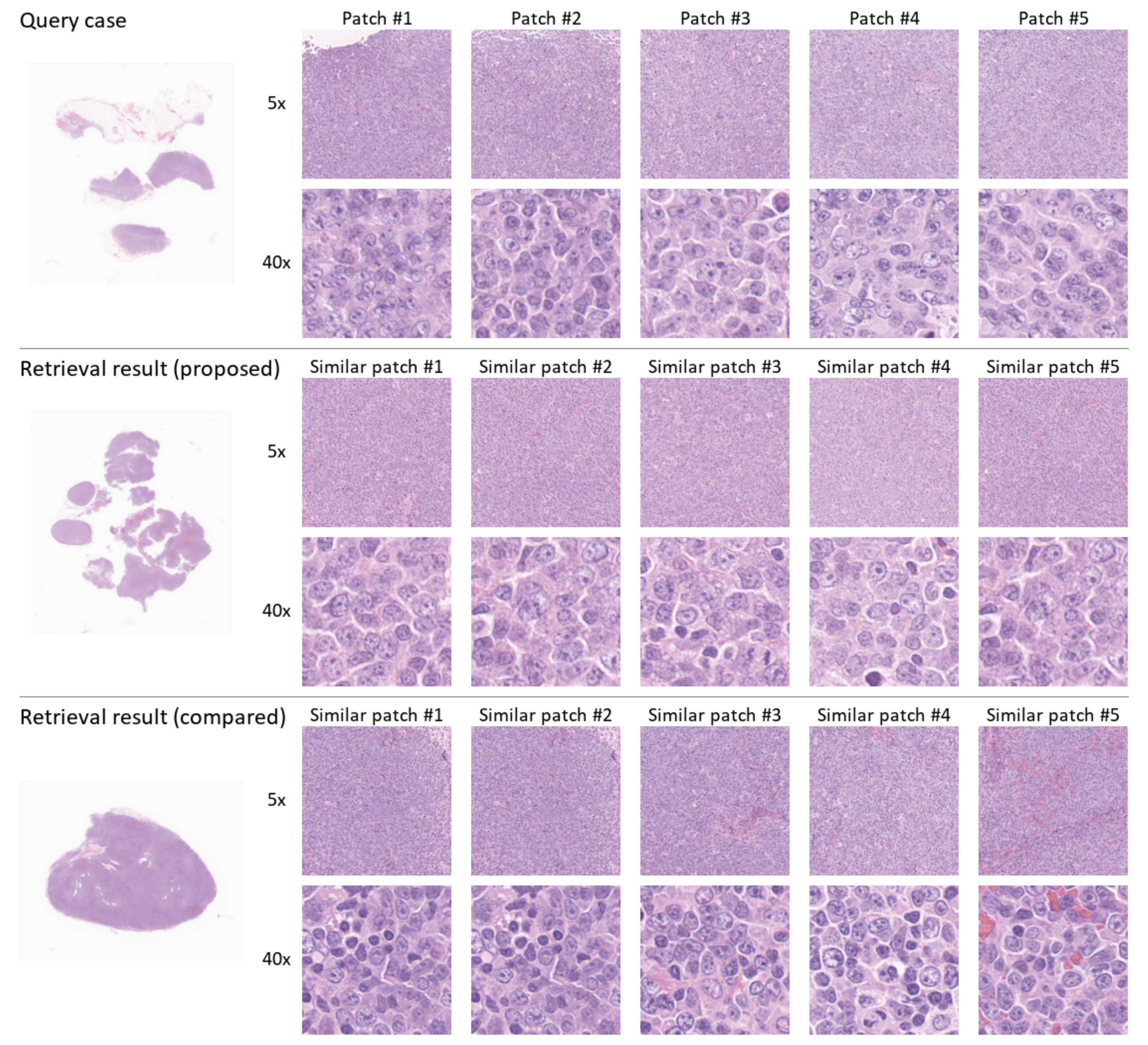}
        }
    \end{minipage}
    &~~
    \begin{minipage}[b]{0.40\hsize}
      \centering
        \fbox{
      \includegraphics[width=\hsize]{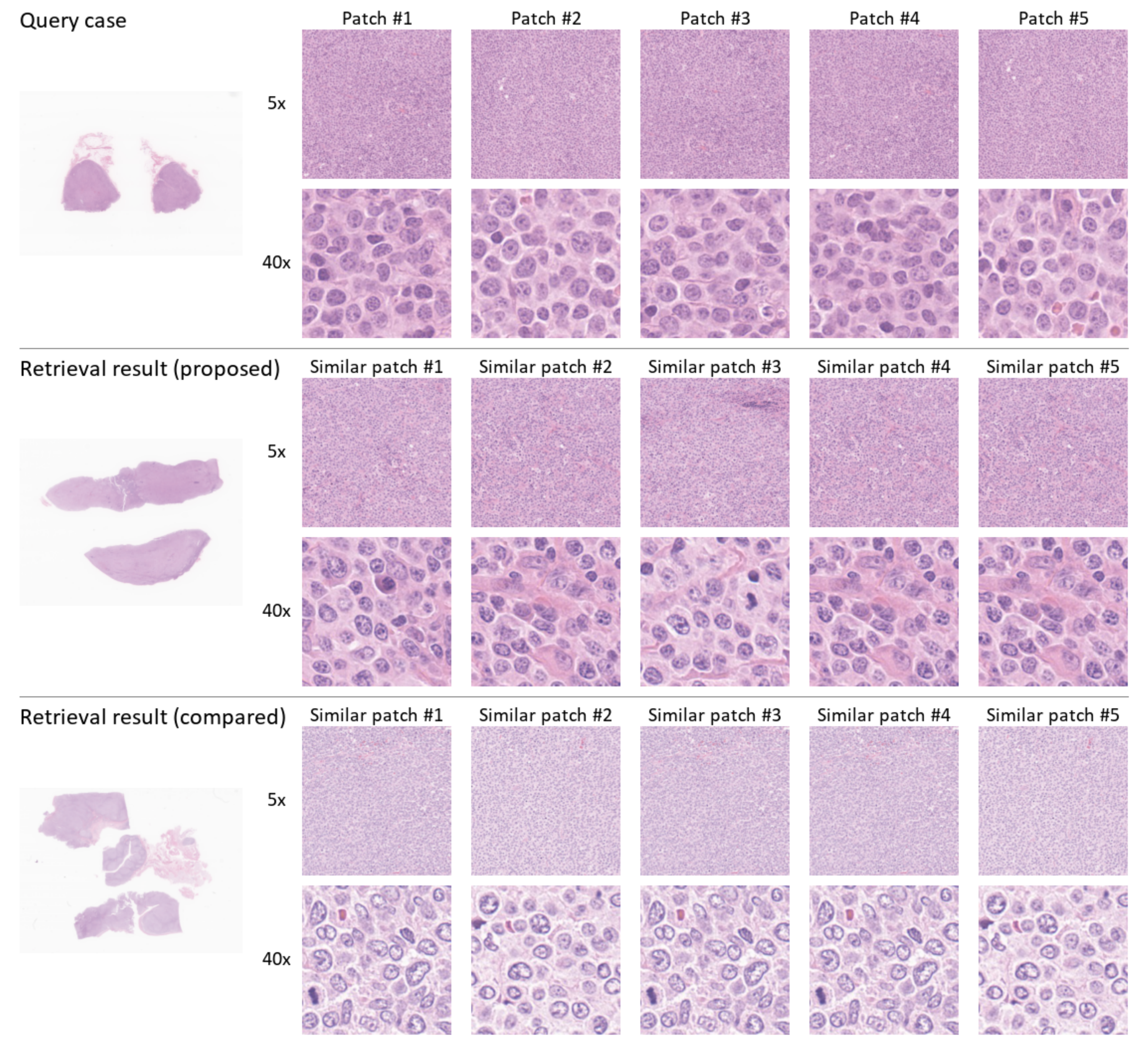}
      }
    \end{minipage}
  \end{tabular}
  \\
  \vspace{2mm}
  \begin{tabular}{cc}
    \begin{minipage}[b]{0.40\hsize}
      \centering
        \fbox{
      \includegraphics[width=\hsize]{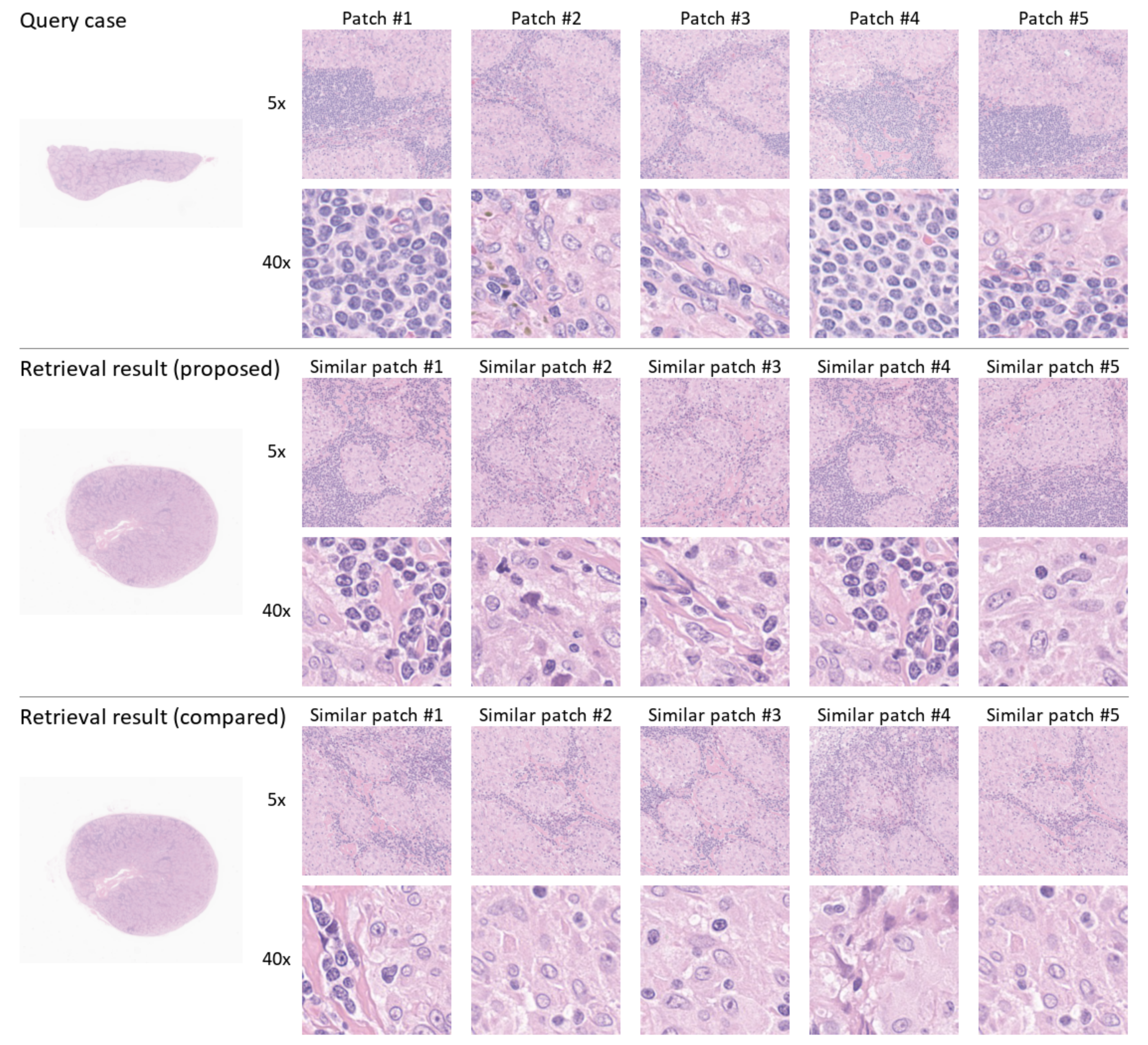}
      }
    \end{minipage}
    &~~
    \begin{minipage}[b]{0.40\hsize}
      \centering
        \fbox{
      \includegraphics[width=\hsize]{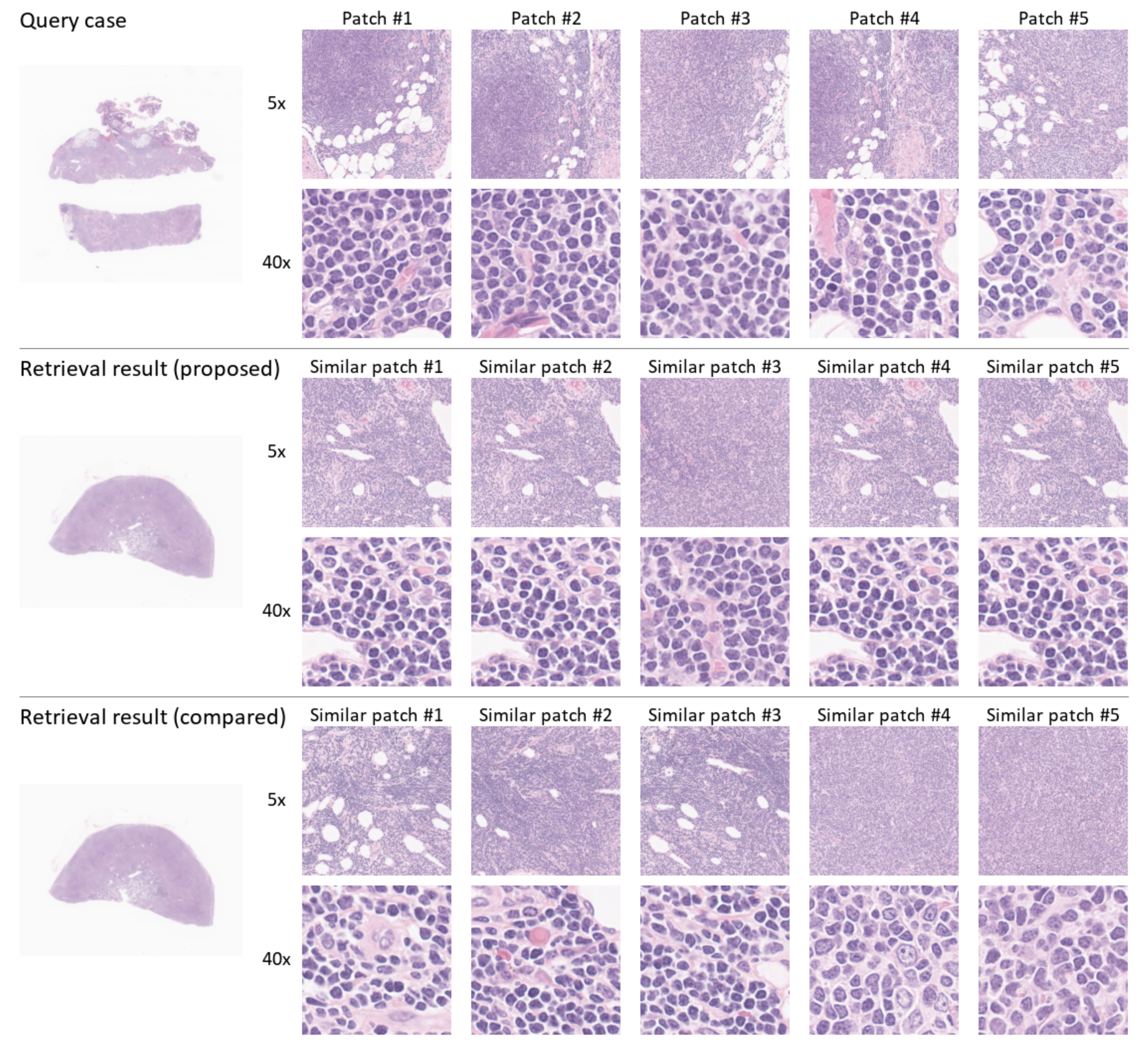}
      }
    \end{minipage}
  \end{tabular}
\end{center}
  \caption{
  Examples of retrieval results that all 10 participants evaluated the proposed method as more similar than the baseline method.
  In addition to the image patches that were shown in the subjective evaluation, the thumbnails of similar cases are also shown.
  As shown in the two bottom examples, even when both methods retrieved the same case, the image patches selected using the proposed SIR method were evaluated as more relevant to the HA patches in the query case.
  }
  \label{fig:res}
\end{figure*}

Figure~\ref{fig:res} shows examples of retrieval results where \emph{all participants} evaluated the proposed method as more similar than the baseline method.
In addition to the same image patches as shown in the subjective evaluation, the thumbnails of the retrieved similar cases are also shown to make it easy to confirm whether the two retrieval cases are the same.
In the examples, the lower images show that both the proposed and baseline methods showed the same similar case (but different image patches).
Even if both methods retrieved the same similar case, the proposed method could obtain more similar image patches and obtain a better evaluation by all 10 pathologists.

\section{Conclusion}
We proposed a case-based SIR method for unannotated large histopathological images of malignant lymphoma.
The proposed method with attention-based MIL can automatically extract informative image patches from unannotated WSIs, and it enables a user to input a WSI as a query without the selection of an image patch.
Moreover, we employed the similarity of IHC staining patterns as the similarity measure in contrastive DML, where the embedded features of the images that have similar IHC staining patterns are much closer.
In the quantitative evaluation of 249 malignant lymphoma patients, we compared the proposed method with several baseline methods, and our proposed method exhibited the highest accuracy in both IHC staining patterns and subtypes between query and similar cases.
Furthermore, we conducted a subjective evaluation experiment to verify our proposed similarity measure using IHC staining patterns and confirmed that our method could retrieve similar cases in which pathologists felt more similar in the observation of the H\&E stained tissue slide than the baseline method.
The proposed case-based SIR method is useful in malignant lymphoma pathology because it provides not only WSIs but also image patches and visualized attention weights that indicate the similarity of the image patches between a query case and a retrieved similar case and the regions of the entire WSI that were focused in the retrieval phase.

\section*{Acknowledgments}
This work was partially supported by MEXT KAKENHI (20H00601, 16H06538, 18H03262), JST CREST (JPMJCR21D3), JST Moonshot R\&D (JPMJMS2033-05), NEDO (JPNP18002, JPNP20006) and RIKEN Center for Advanced Intelligence Project.

\clearpage

{\small
\bibliographystyle{ieee}
\bibliography{ref}
}

\end{document}